\documentclass{article}

\PassOptionsToPackage{numbers,sort&compress}{natbib}

\usepackage[eandd,preprint]{neurips_2026}

\usepackage[utf8]{inputenc}
\usepackage[T1]{fontenc}
\usepackage{hyperref}
\usepackage{url}
\usepackage{booktabs}
\usepackage{amsfonts}
\usepackage{amsmath}
\usepackage{nicefrac}
\usepackage{microtype}
\usepackage{xcolor}
\usepackage{graphicx}
\usepackage{multirow}
\usepackage{float}
\usepackage{tcolorbox}
\usepackage{tabularx}

\newcommand{\benchmark}{FOCUS}
\newcommand{\benchmarklong}{Foundation Ophthalmic Cross-Dataset Understanding under Shift}
\graphicspath{{figures/}{images/}}

\title{\benchmark: Benchmarking Retinal Model Generalization from Foundation Vision Encoders to Multimodal LLMs}

\author{%
\parbox{0.95\textwidth}{%
\centering
David Restrepo\textsuperscript{1,2} \quad
Chenwei Wu\textsuperscript{3} \quad
Luis Filipe Nakayama\textsuperscript{4} \quad
Miguel L. Martins\textsuperscript{5,6}
\\[0.4ex]
Stergios Christodoulidis\textsuperscript{1,2} \quad
Maria Vakalopoulou\textsuperscript{1,2} \quad
Enzo Ferrante\textsuperscript{7,8}
\\[1.2ex]
\textsuperscript{1}CentraleSupélec, Université Paris-Saclay, France
\\[0.3ex]
\textsuperscript{2}Cancer Data Science Unit, IHU PRISM,
Université Paris-Saclay, CentraleSupélec,\\
Gustave Roussy, INSERM, France
\\[0.3ex]
\textsuperscript{3}University of Michigan, USA
\quad
\textsuperscript{4}Federal University of São Paulo, Brazil
\quad
\textsuperscript{5}University of Porto, Portugal
\quad
\textsuperscript{6}Aalborg University, Denmark
\quad
\textsuperscript{7}Instituto de ciencias de la computaci\'on, CONICET - Universidad de Buenos Aires, Argentina
\quad
\textsuperscript{8}Instituto Tecnológico de Buenos Aires (ITBA), Argentina
}}

\begin{document}

\maketitle

\begin{abstract}
Progress in AI-based retinal image analysis has accelerated with the adoption of foundation models, yet evaluating their real-world reliability remains challenging. Performance reported on a single dataset does not capture how models behave under dataset shift, across clinical definitions, or for different patient subgroups. This limitation is particularly critical in medical imaging analysis, where robustness, calibration, and fairness are essential for safe deployment. We introduce \benchmark{} (\benchmarklong), a cross-dataset benchmark for evaluating retinal fundus models that considers vision-only encoder models (VM), vision-language dual-encoder models (VLM), and multimodal large language models (MLLM). \benchmark{} harmonizes binary diabetic retinopathy, referable diabetic retinopathy, and glaucomatous optic neuropathy tasks across ten public datasets spanning diverse geographies, acquisition conditions, and label protocols. The benchmark evaluates models through a unified analysis layer that measures ranking performance, calibration, subgroup disparities, and image-quality robustness. We present a large-scale evaluation covering 532 base configurations and 228 MLLM configurations adapted through supervised fine-tuning with
low-rank adaptation (LoRA). Results show that no model family consistently dominates across tasks and datasets: general VM encoders achieve the strongest average ranking performance, medical MLLMs are competitive but variable, and dual encoder VLMs benefit substantially from lightweight adaptation. Fine-tuning improves in-domain performance but exhibits heterogeneous transfer to external datasets, particularly in calibration. These findings demonstrate that retinal model evaluation is inherently multidimensional. \benchmark{} provides a practical framework and public benchmark to assess generalization, reliability, and robustness beyond single-dataset leaderboards.
\end{abstract}

\section{Introduction}
 
Retinal fundus photography is one of the most widely used imaging modalities for large-scale screening and monitoring for vision-threatening conditions \citep{fundusscreening}. However, it presents a unique challenge for foundation-model evaluation and training due to the confluence of dataset shift and label noise \citep{messidor2_grades2018}. Beyond technical variations in acquisition (e.g., field of view, illumination), annotated labels exhibit significant inter-grader variability and differing levels of clinical expertise \citep{messidor2_grades2018}. Foundation models trained on large datasets often overfit to overrepresented populations or specific acquisition hardware, creating hidden biases that limit clinical utility and fairness when deployed in diverse, real-world healthcare systems \citep{bias_marzyeh,bias_oph,fair_med_found}.

The evaluation problem has become more urgent as AI-based retinal image analysis has moved beyond task-specific classifiers. Current systems include foundation vision-only encoder models (VMs) \citep{retfound2023,visionfm2023}, dual encoder-only vision-language models (VLMs)\citep{fair_med_found,eyeclip2025,retclip2024}, and generative instruction-tuned multimodal large language models (MLLMs)\citep{medgemma15_2026,medgemma2025}. These systems expose different prediction interfaces: the VM and VLM encoders generate embeddings and multi-modal embeddings, and the MLLMs return token logits to be used in an auto-regressive way. Without a shared evaluation layer, apparent gains can reflect changes in prompt template, threshold, calibration convention, or label mapping rather than real clinical transfer.
    
Existing resources cover important pieces of this problem, but not the full benchmark need. Single datasets and challenges such as IDRiD \citep{idrid2018}, RFMiD \citep{rfmid2_2023}, BRSET \citep{brset_plos2024}, mBRSET \citep{mbrset_scidata}, PAPILA \citep{papila2022}, and G1020 \citep{g1020_2020} provide valuable disease labels and sometimes demographic or image-quality metadata, but they do not by themselves test cross-dataset transfer across model families (VM, VLM-encoders, MLLMs). Recent MLLM and LVLM benchmarks, such as FunBench \citep{funbench2025} and LMOD \citep{lmod2025}, evaluate ophthalmic reasoning, but they are not designed as a harmonized evaluation suite for measuring disease-classification transfer across the full architectural spectrum covering VMs, VLMs, and MLLMs systems together. Table~\ref{tab:related-benchmarks} summarizes this gap: prior resources usually cover either diseases, metadata, or generative models, while few combine evaluations on some of the Diabetic Retinopathy (DR), referable DR, glaucoma, out-of-domain evaluation, fairness, quality robustness, VLMs, MLLMs, and fine-tuning transfer in one framework.

In this study, we present \benchmark{}, a benchmark for comparing foundation models from vision-only encoders to generative multimodal models. The benchmark asks a deliberately practical question: when foundation models are adapted to domain-specific fundus data, do they improve the target task only on the training dataset, or do they also preserve ranking performance, calibration, subgroup behavior, and image-quality robustness? By combining ten fundus datasets across multiple regions and acquisition conditions, \benchmark{} creates a broader test set for out-of-distribution tasks and demographic variation than any single dataset can provide.

Our contributions are: (1) a dataset registry and task harmonization layer for binary diabetic retinopathy, referable diabetic retinopathy, and glaucomatous optic neuropathy across ten fundus datasets; (2) a unified evaluation layer covering 532 model--dataset-task-method configurations, including 190 for VMs, 228 for VLMs, and 114 for MLLMs; (3) an evaluation across four axes: performance, calibration, fairness, and image-quality robustness; and (4) a Supervised Fine Tuning (SFT) Low-rank adaptation (LoRA) analysis for MLLMs for the four evaluation axes, comparing each adapter with its matching base model on same-dataset test sets and same-task external datasets.

We release the code and modular pipelines for benchmark evaluation and dataset downloading/cleaning through a \href{https://github.com/dsrestrepo/FOCUS-Retinal-Benchmark}{repository}, together with an interactive \href{https://huggingface.co/spaces/dsrestrepo/FOCUS-Retinal-Benchmark}{visualization dashboard}.

\section{Related Work}

\paragraph{VMs and VLMs for fundus imaging.}
Retinal AI has moved from task-specific "specialist" models toward foundation models that differ in domain pretraining and output interface. \benchmark{} therefore keeps VM, VLM, and MLLM interfaces separate while evaluating them through one harmonized retinal analysis layer. General VM/VLM backbones such as ViT~\citep{vit2021}, DINOv2~\citep{dinov2_2023}, DINOv3~\citep{dinov3_2025}, CLIP~\citep{clip2021}, and SigLIP2~\citep{siglip2_2025} provide broad visual or image-text representations. Medical and ophthalmic variants adapt these ideas to medical images: RETFound~\citep{retfound2023} and VisionFM~\citep{visionfm2023} specialize VM pretraining for ophthalmology, MedSigLIP~\citep{medgemma2025} adapts dual-encoder image-text modeling to medical images, and FLAIR~\citep{flair2025}, RET-CLIP~\citep{retclip2024}, and EyeCLIP~\citep{eyeclip2025} focus on retinal or ophthalmic VLM pretraining.

\paragraph{MLLMs for fundus imaging.}
Generative MLLMs add a prompt-based interface with text outputs and token probabilities. General models such as Gemma-3~\citep{gemma3_2025}, Qwen3-VL~\citep{qwen3vl2025}, and LLaVA-NeXT~\citep{llavanext2024} provide broad image-language reasoning, while medical or ophthalmic models such as LLaVA-Med~\citep{llavamed2023}, Med-Flamingo~\citep{medflamingo2023}, MedGemma~\citep{medgemma2025}, MedGemma 1.5~\citep{medgemma15_2026}, and OphGLM~\citep{ophglm2023} adapt generative VLMs using biomedical captions, medical instruction tuning, or ophthalmology chat data.

\paragraph{Benchmarking retinal and ophthalmic models.}

Public fundus datasets and challenges have been essential for diabetic retinopathy, referable diabetic retinopathy, glaucoma, and multi-disease recognition, including BRSET\citep{brset_physionet2024}, mBRSET\citep{mbrset_scidata,mbrset_physionet2024}, PAPILA\citep{papila2022}, RFMiD\citep{rfmid2021}, RFMiD 2.0\citep{rfmid2_2023}, IDRiD\citep{idrid2018}, Messidor-2\citep{messidor2_grades2018}, G1020\citep{g1020_2020}, REFUGE\citep{refuge}, and JSIEC1000~\citep{jsiec1000_kaggle}. It is common for methodological studies to conduct their experiments on a limited set of datasets, limited or heterogeneous model interfaces and downstream performance metrics. Ophthalmological benchmarks have thus been proposed to evaluate models across methodologies in a unified way. FunBench evaluates fundus-reading skills of MLLMs with vision-question answering (VQA) tasks~\citep{funbench2025}; LMOD focuses on measuring large-scale ophthalmological VLM vulnerabilities, such as  hallucinations~\citep{lmod2025}; MultiEYE and MIRAGE emphasize multimodal or optical coherence tomography (OCT)-centered retinal analysis~\citep{multieye2025,mirage2025}. While these provide valuable insights into specific failure modes, they often focus on a single model paradigm (e.g., generative models) or lack a harmonized framework for cross-dataset disease transfer.

\begin{figure*}[t]
  \centering
  \includegraphics[width=\linewidth]{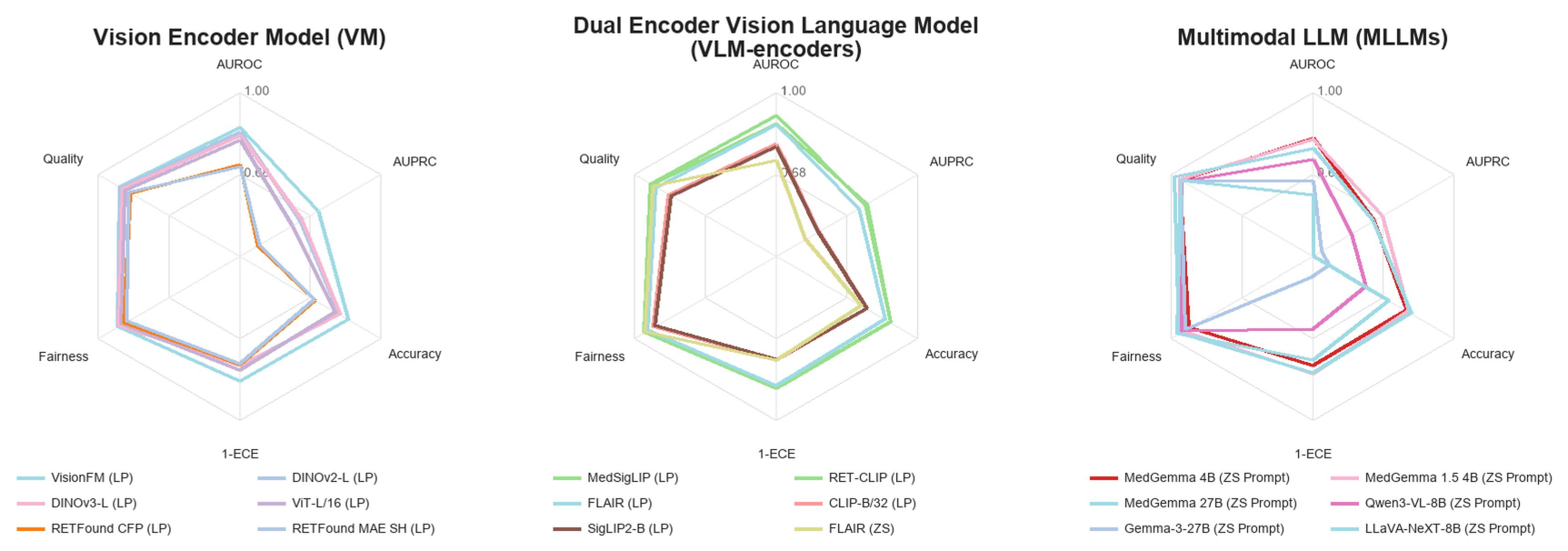}
  \caption{\textbf{Multimetric arena view.} Spider plots summarize the top 6 VM, VLM-encoders, and MLLM configurations across AUROC, AUPRC, accuracy, calibration, fairness diagnostics, and image-quality robustness. Colors distinguish models; LP = Linear probing, ZS = Zero-shot, ZS Prompt = Zero-shot prompting. This view makes visible that strong ranking performance need not imply strong calibration or balanced subgroup/quality behavior.}
  \label{fig:arena-spider}
\end{figure*}

\paragraph{The proposed benchmark.} \benchmark{} was created specifically for colored fundus photographs and it is designed to assess whether the foundation models generalize beyond their training or fine-tuning datasets. As a whole, the data in \benchmark{} is diverse in terms of acquisition conditions, disease tasks and demographic variables (e.g., age, gender, geography). Specifically, it provides a plug-and-play evaluation tool covering a wide range of population and acquisition shifts for same-task out-of-domain evaluation, while  simultaneously harmonizing the available metadata to facilitate analysis of prediction calibration, age/sex subgroup gaps, and image-quality robustness. Table~\ref{tab:related-benchmarks} summarizes this gap.

\begin{table*}[t]
  \caption{Coverage comparison with related fundus datasets, challenges, and ophthalmic benchmarks.}
  \label{tab:related-benchmarks}
  \centering
  \scriptsize
  \setlength{\tabcolsep}{1.8pt}
  \renewcommand{\arraystretch}{1.03}
  \begin{tabularx}{\linewidth}{@{}>{\raggedright\arraybackslash}p{0.20\linewidth}>{\raggedright\arraybackslash}X*{3}{>{\centering\arraybackslash}p{0.039\linewidth}}>{\centering\arraybackslash}p{0.057\linewidth}*{5}{>{\centering\arraybackslash}p{0.039\linewidth}}>{\centering\arraybackslash}p{0.050\linewidth}@{}}
    \toprule
    Resource & Type & DR & \shortstack{Ref.\\DR} & Glauc. & \shortstack{multi-\\reg.} & Fair. & Qual. & VM & VLM & MLLM & \shortstack{FT\\trans.} \\
    \midrule
    IDRiD~\citep{idrid2018} & DR challenge
      & Y & Y & -- & -- & -- & -- & D & -- & -- & -- \\
    RFMiD / RFMiD 2.0~\citep{rfmid2021,rfmid2_2023} & Multi-disease fundus datasets
      & Y & D & D & -- & -- & -- & D & -- & -- & -- \\
    BRSET / mBRSET~\citep{brset_plos2024,brset_physionet2024,mbrset_physionet2024,mbrset_scidata} & Brazilian fundus datasets
      & Y & Y & Y & D & D & D & D & -- & -- & -- \\
    PAPILA / G1020 / REFUGE~\citep{papila2022,g1020_2020,refuge} & Glaucoma datasets/challenges
      & -- & -- & Y & D & D & -- & D & -- & -- & -- \\
    FunBench~\citep{funbench2025} & Fundus MLLM VQA benchmark
      & VQA & VQA & VQA & Y & -- & -- & -- & -- & Y & -- \\
    LMOD~\citep{lmod2025} & Ophthalmology LVLM benchmark
      & VQA & VQA & VQA & Y & D & -- & -- & D & Y & -- \\
    MultiEYE / MIRAGE~\citep{multieye2025,mirage2025} & Multimodal/OCT retinal benchmarks
      & D & -- & D & Y & -- & -- & Y & D & D & -- \\
    \benchmark{} & Cross-dataset fundus benchmark
      & Y & Y & Y & Y & Y & Y & Y & Y & Y & Y \\
    \bottomrule
  \end{tabularx}
  \vspace{0.25em}
  {\tiny Y: explicit benchmark evaluation; D: labels/metadata available but not a cross-model benchmark axis; VQA: broader visual-question-answering or MLLM evaluation; --: not central to the resource.}
\end{table*}

\section{\benchmark{} Benchmark Design}

\subsection{Datasets and tasks}

\benchmark{} targets color fundus photography datasets with complementary metadata relating to the acquisition conditions, demographics, and a plurality of ophthalmological downstream tasks. The current dataset registry includes BRSET~\citep{brset_plos2024,brset_physionet2024}, mBRSET~\citep{mbrset_physionet2024,mbrset_scidata}, PAPILA~\citep{papila2022}, RFMiD~\citep{rfmid2021}, RFMiD 2.0~\citep{rfmid2_2023}, IDRiD~\citep{idrid2018}, Messidor-2 labels~\citep{messidor2_grades2018}, REFUGE~\citep{refuge}, G1020~\citep{g1020_2020}, and JSIEC1000~\citep{jsiec1000_kaggle}. Not all datasets are uniformly annotated, thus task-specific analyses are conditioned only on the subset of data containing valid ground truth annotations. An overview of the datasets can be seen in Appendix~\ref{apendix:datasets}.

\paragraph{Task harmonization.}
\label{sec:task-harmonization}
\benchmark{} harmonizes dataset-specific annotations into three binary clinical tasks: \textit{any} diabetic retinopathy (DR), \textit{referable} DR, and \textit{glaucomatous optic neuropathy} (GON). Any DR captures early pathological signs using ICDR grade $\ge 1$~\citep{icdr}, while referable DR reflects the screening decision to refer moderate non-proliferative DR or worse (ICDR grade $\ge 2$) and/or suspicious macular edema. We use GON rather than definitive glaucoma because fundus photography captures structural risk markers, such as neuroretinal rim thinning or increased vertical cup-to-disc ratio; labels are harmonized using VCDR $\ge 0.6$ or specialist annotation of glaucomatous damage~\citep{gon}. Together, these tasks span localized vascular lesions, referral-level disease, and global optic-disc structure.

Each model--dataset--task--method configuration is evaluated separately; images are not pooled across datasets into one training or test population. Harmonization defines common evaluation endpoints but does not make source grading criteria equivalent, particularly for GON. Dataset-specific annotation provenance and operational mappings are provided in Appendix~\ref{app:label-provenance}.

\subsection{Model evaluation}

This benchmark separates two axes that are often conflated: domain specificity and output interface (see Table \ref{tab:model-families}). VMs are evaluated through frozen-feature linear probes, including retinal foundation encoders such as RETFound~\citep{retfound2023}. VLMs are evaluated through zero-shot image-text embedding matching and frozen-image-feature linear probing. MLLMs are evaluated prompting the model to return binary outputs ('yes' or 'no'), and then using the first-token positive-class probabilities (see Appendix~\ref{llm:inference} for details). Primary MLLM metrics use raw first-token affirmative probabilities and decisions parsed from generated answers.

Within each dataset-task pair, pipelines use the same test split and harmonized targets across all the experiments. Calibration and decision metrics reflect interface choices as well as model behavior: logistic probe probabilities, scaled image-text similarities, and affirmative-token probabilities have different constructions. Within-interface type comparisons provide more controlled comparisons of models, while cross-interface comparisons characterize alternative ways of solving the same task.

We additionally assess sensitivity to linear-probe random initialization, encoder-VLM and MLLM prompt wording, MLLM response parsing, and LoRA configuration. The experimental protocols and detailed results are reported in Appendix~\ref{app:sensitivity}.

\begin{table}[t]
  \caption{Model families represented in the current \benchmark{} result bundle.}
  \label{tab:model-families}
  \centering
  \scriptsize
  \setlength{\tabcolsep}{3pt}
  \renewcommand{\arraystretch}{0.92}
  \begin{tabularx}{0.82\linewidth}{@{}ll>{\raggedright\arraybackslash}X>{\raggedright\arraybackslash}p{0.25\linewidth}@{}}
    \toprule
    Family & Domain & Architectures & Interface \\
    \midrule
    VM & Generalist & ViT-L/16, DINOv2-L, DINOv3-ViT-L & Image embeddings \\
    VM & Ophthalmic & RETFound variants, VisionFM-Fundus & Image embeddings \\
    \midrule
    VLM & Generalist & CLIP, SigLIP2, MedSigLIP & Joint image-text embeddings \\
    VLM & Ophthalmic & EyeCLIP, RET-CLIP, FLAIR & Joint image-text embeddings \\
    \midrule
    MLLM & Generalist & Qwen3-VL, Gemma-3, LLaVA-NeXT & Text output and token logits \\
    MLLM & Medical & MedGemma variants & Text output and token logits \\
    \bottomrule
  \end{tabularx}
\end{table}

\subsection{Evaluation axes}

\benchmark{} is organized around four primary evaluation axes. \emph{Performance} is assessed using threshold-free metrics (e.g. AUROC) that summarize how well model scores separate positive from negative cases across operating points, together with threshold-dependent metrics that reflect classification quality at a fixed decision threshold (e.g. Accuracy). \emph{Calibration} measures whether predicted positive-class scores correspond to the observed frequency of positive classes for each harmonized task. 
\emph{Fairness} summarizes differences in performance across age and sex subgroups (when the required metadata is available).
\emph{Robustness} evaluates whether performance and calibration remain stable across image-quality strata for datasets with quality annotations. Image quality is taken from the dataset-provided quality metadata when available, typically reflecting acquisition or grading quality labels assigned during dataset curation.

These axes address complementary concerns in retinal screening: ranking performance measures whether disease-positive cases can be prioritized for specialist assessment; calibration measures whether reported confidence agrees with observed disease frequency; subgroup analyses identify populations with disproportionate error rates; and image-quality analyses assess sensitivity to heterogeneous acquisition conditions. Demographic analyses are restricted to BRSET, mBRSET, and PAPILA where the required metadata, task labels, and subgroup sizes permit evaluation. Image-quality analyses are restricted to BRSET and mBRSET. Consequently, these analyses have narrower coverage than the primary performance evaluation (Table~\ref{tab:datasets}).

Fine-tuned MLLMs are evaluated separately from the base benchmark. We train SFT LoRA adapters only on BRSET and mBRSET, because these are the configured training datasets that support all three harmonized tasks and include the row-level metadata needed for performance, calibration, fairness, and image-quality robustness analyses. For each model--training dataset--task combination, we measure in-dataset performance on the held-out test split of the training dataset and same-task out-of-distribution (OOD) transfer on every other benchmark dataset that supports the task. Results are reported as metric deltas relative to the matching base MLLM evaluated on the same test dataset and task.

\begin{figure*}[t]
  \centering
  \includegraphics[width=\linewidth]{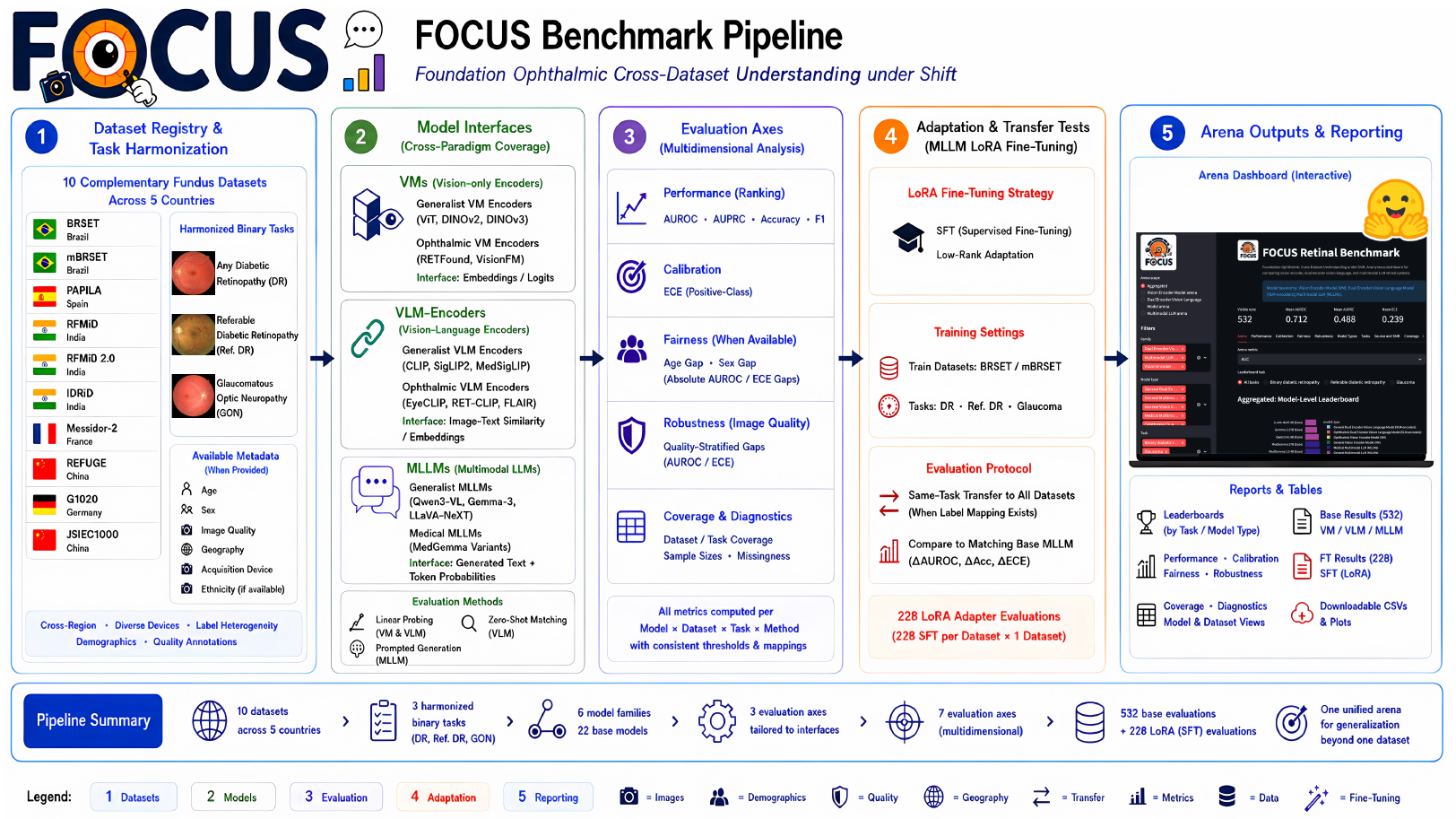}
  \caption{\textbf{Benchmark pipeline.} \benchmark{} connects dataset/task harmonization, model-interface-specific inference, multidimensional evaluation, LoRA adaptation tests, and arena-style reporting. The intent is to measure whether fundus-specific adaptation generalizes rather than only improving the dataset used for training.}
  \label{fig:benchmark-overview}
\end{figure*}

\section{Empirical Results}

The base benchmark suite covers 532 model--dataset--task--method configurations: of those 190 correspond to VMs, 228 to VLMs, and 114 to MLLMs. The language-model fine-tuning suite covers 228 SFT adapter evaluations and their full details are in Appendices \ref{expdesign} and \ref{llm:ft}. Tab.\ref{tab:model-type-results} shows the method-average results which we will unpack in the rest of this section : across four axes of performance, calibration, fairness and robustness, no single model or model class dominates. 

\begin{tcolorbox}[colback=blue!5, colframe=blue!30, boxrule=0.5pt, left=4pt, right=4pt, top=3pt, bottom=3pt]
\small\textbf{Finding 1: Generalist pretrained VM encoder remains a surprisingly strong and robust retinal baseline.} 
\end{tcolorbox}

General pretrained VM encoders achieve the strongest mean AUROC in FOCUS, surpassing ophthalmic-domain VM encoders across most datasets. As shown in Tab. \ref{tab:per_dataset_cv_gap}, the largest gap appears on mBRSET (0.162), the only mobile-phone fundus dataset in the benchmark. This suggests that current ophthalmic VM pretraining may be less robust to acquisition shifts outside conventional desktop fundus photography. Because mBRSET also differs in population and label composition, this gap should not be interpreted as isolating device shift alone; rather, it highlights a deployment-relevant setting where ophthalmic-domain pretraining underperforms generalist visual pretraining.  Strong generalist visual representations remain difficult baselines to beat, especially under cross-dataset and device-shift evaluation. As for other model groups, VLM performance depends on the evaluation interface: Ophthalmic VLMs outperform general VLMs on mean AUROC in zero-shot settings but fall short in linear probing, showing lower performance ceiling.
 Lastly, medical MLLMs substantially outperform general MLLMs.

\begin{table*}[t]
  \caption{Method-aware mean metrics by model type and task. Higher is better except for ECE.}
  \label{tab:model-type-results}
  \centering
  \scriptsize
  \setlength{\tabcolsep}{5pt}
  \begin{tabular}{lccccccccc}
    \toprule
    Task & Model type & Method & Accuracy & AUROC & AUPRC & ECE & Fairness & Quality & Shift \\
    \midrule
    \multicolumn{10}{l}{\textbf{Binary DR}} \\
    \quad  & General VM & Linear probing & 0.813 & 0.867 & 0.782 & 0.188 & 0.932 & 0.913 & 0.739 \\
    \quad  & Ophthalmic VM & Linear probing & 0.720 & 0.690 & 0.537 & 0.206 & 0.922 & 0.910 & 0.703 \\
    \quad  & General VLM-encoders & Linear probing & \textbf{0.819} & 0.865 & \textbf{0.784} & 0.179 & 0.936 & 0.874 & 0.746 \\
    \quad  & General VLM-encoders & Zero-shot & 0.477 & 0.547 & 0.346 & 0.311 & 0.948 & 0.897 & 0.840 \\
    \quad  & Ophthalmic VLM-encoders & Linear probing & 0.780 & 0.814 & 0.702 & \textbf{0.151} & 0.916 & 0.911 & 0.801 \\
    \quad  & Ophthalmic VLM-encoders & Zero-shot & 0.747 & 0.698 & 0.522 & 0.215 & \textbf{0.954} & 0.900 & \textbf{0.845} \\
    \quad  & General MLLMs & Zero-shot prompting & 0.521 & 0.685 & 0.457 & 0.435 & 0.948 & 0.911 & 0.742 \\
    \quad  & Medical MLLMs & Zero-shot prompting & 0.791 & \textbf{0.868} & 0.707 & 0.194 & 0.936 & \textbf{0.939} & 0.820 \\
    \midrule
    \multicolumn{10}{l}{\textbf{Referable DR}} \\
    \quad  & General VM & Linear probing & \textbf{0.858} & 0.850 & 0.556 & 0.194 & 0.907 & 0.857 & 0.741 \\
    \quad  & Ophthalmic VM & Linear probing & 0.765 & 0.697 & 0.363 & 0.233 & 0.913 & 0.916 & 0.799 \\
    \quad  & General VLM-encoders & Linear probing & 0.845 & \textbf{0.853} & 0.544 & 0.189 & 0.908 & 0.839 & \textbf{0.860} \\
    \quad  & General VLM-encoders & Zero-shot & 0.654 & 0.596 & 0.314 & 0.209 & 0.940 & 0.907 & 0.817 \\
    \quad  & Ophthalmic VLM-encoders & Linear probing & 0.776 & 0.770 & 0.528 & \textbf{0.178} & 0.910 & 0.903 & 0.674 \\
    \quad  & Ophthalmic VLM-encoders & Zero-shot & 0.440 & 0.672 & 0.360 & 0.322 & \textbf{0.952} & 0.930 & 0.739 \\
    \quad  & General MLLMs & Zero-shot prompting & 0.517 & 0.651 & 0.393 & 0.433 & 0.932 & 0.940 & 0.446 \\
    \quad  & Medical MLLMs & Zero-shot prompting & 0.775 & 0.844 & \textbf{0.700} & 0.208 & 0.916 & \textbf{0.959} & 0.389 \\
    \midrule
    \multicolumn{10}{l}{\textbf{Glaucoma}} \\
    \quad  & General VM & Linear probing & 0.712 & 0.742 & 0.471 & 0.259 & 0.872 & 0.884 & 0.747 \\
    \quad  & Ophthalmic VM & Linear probing & 0.707 & 0.591 & 0.312 & 0.263 & 0.915 & 0.872 & \textbf{0.903} \\
    \quad  & General VLM-encoders & Linear probing & 0.717 & \textbf{0.745} & 0.474 & 0.248 & 0.924 & 0.875 & 0.771 \\
    \quad  & General VLM-encoders & Zero-shot & 0.516 & 0.488 & 0.226 & 0.317 & 0.921 & 0.869 & 0.880 \\
    \quad  & Ophthalmic VLM-encoders & Linear probing & 0.721 & 0.719 & \textbf{0.527} & 0.230 & 0.919 & 0.919 & 0.806 \\
    \quad  & Ophthalmic VLM-encoders & Zero-shot & 0.757 & 0.541 & 0.266 & \textbf{0.162} & 0.940 & 0.924 & 0.878 \\
    \quad  & General MLLMs & Zero-shot prompting & 0.730 & 0.630 & 0.418 & 0.259 & \textbf{0.964} & \textbf{1.000} & 0.489 \\
    \quad  & Medical MLLMs & Zero-shot prompting & \textbf{0.816} & 0.656 & 0.470 & 0.176 & 0.957 & 1.000 & 0.482 \\
    \midrule
    \multicolumn{10}{l}{\textbf{Overall}} \\
    \quad  & General VM & Linear probing & \textbf{0.803} & 0.828 & 0.617 & 0.209 & 0.899 & 0.885 & 0.654 \\
    \quad  & Ophthalmic VM & Linear probing & 0.733 & 0.667 & 0.414 & 0.231 & 0.917 & 0.899 & 0.762 \\
    \quad  & General VLM-encoders & Linear probing & 0.802 & \textbf{0.829} & 0.614 & 0.201 & 0.923 & 0.863 & 0.661 \\
    \quad  & General VLM-encoders & Zero-shot & 0.552 & 0.549 & 0.303 & 0.275 & 0.934 & 0.891 & \textbf{0.800} \\
    \quad  & Ophthalmic VLM-encoders & Linear probing & 0.763 & 0.773 & 0.592 & \textbf{0.182} & 0.915 & 0.911 & 0.732 \\
    \quad  & Ophthalmic VLM-encoders & Zero-shot & 0.636 & 0.647 & 0.395 & 0.240 & 0.948 & 0.918 & 0.729 \\
    \quad  & General MLLMs & Zero-shot prompting & 0.575 & 0.658 & 0.423 & 0.388 & \textbf{0.950} & 0.950 & 0.603 \\
    \quad  & Medical MLLMs & Zero-shot prompting & 0.791 & 0.803 & \textbf{0.642} & 0.194 & 0.939 & \textbf{0.966} & 0.567 \\
    \bottomrule
  \end{tabular}
\end{table*}

Additional task-level comparisons and leaderboard views can be found in the Appendix, in
Figures~\ref{fig:task-model-type} and \ref{fig:model-leaderboard}.

\begin{table}[t]
  \centering
  \caption{Per-dataset mean AUROC for VM encoders. $\Delta=\mathrm{AUROC}_{\mathrm{gen}}-\mathrm{AUROC}_{\mathrm{oph}}$.}
  \label{tab:per_dataset_cv_gap}
  \scriptsize
  \setlength{\tabcolsep}{10pt}
  \renewcommand{\arraystretch}{0.92}
  \begin{tabular}{lllccc}
    \toprule
    Dataset & Country & Device & Gen. & Oph. & $\Delta$ \\
    \midrule
    \textbf{mBRSET} & Brazil & \textbf{mobile} & 0.754 & 0.592 & \textbf{+0.162} \\
    IDRiD      & India   & desktop & 0.825 & 0.692 & +0.133 \\
    BRSET      & Brazil  & desktop & 0.822 & 0.700 & +0.122 \\
    JSIEC1000  & China   & desktop & 0.958 & 0.851 & +0.106 \\
    Messidor-2 & France  & desktop & 0.836 & 0.759 & +0.077 \\
    RFMiD      & India   & desktop & 0.914 & 0.849 & +0.066 \\
    PAPILA     & Spain   & desktop & 0.766 & 0.710 & +0.056 \\
    G1020      & Germany & desktop & 0.631 & 0.614 & +0.017 \\
    RFMiD-2    & India   & desktop & 0.846 & 0.918 & -0.072 \\
    \bottomrule
  \end{tabular}
\end{table}

\begin{figure*}[t]
  \centering
  \includegraphics[width=\textwidth]{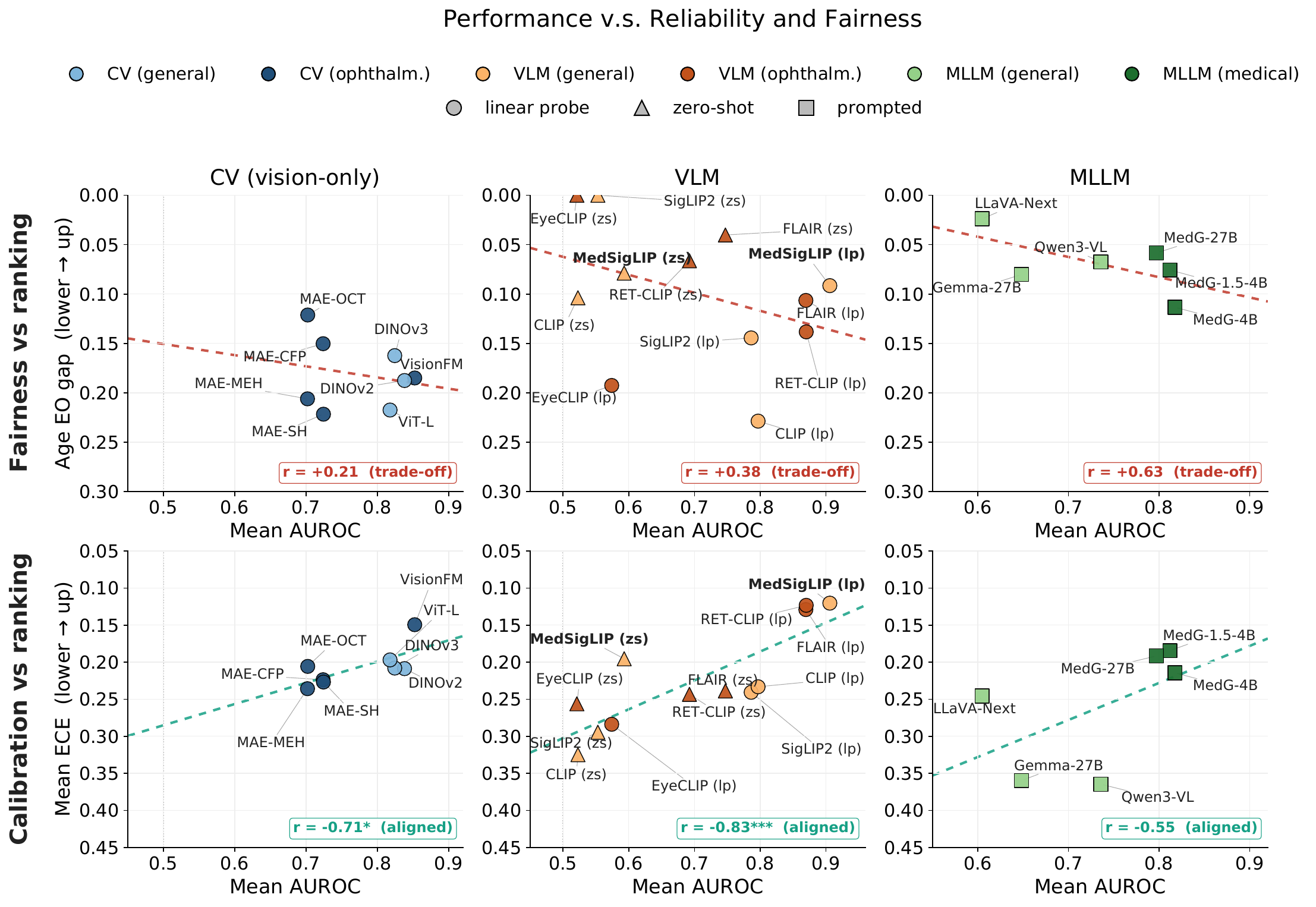}
  \caption{\textbf{Performance vs.\ reliability and fairness, by model family.}
    Each point is one (model, method) configuration; rows index axis
    (top: age equalised-odds gap; bottom: ECE), columns index family
    (CV / VLM / MLLM). Marker colour encodes model type, marker shape
    encodes adaptation method (circle = linear probe, triangle =
    zero-shot, square = prompted). Both axes are
    inverted so that ``up'' is desirable on every panel. The dotted
    line in each panel is a least-squares fit on that panel's points;
    \textcolor[HTML]{C0392B}{\textbf{red}} denotes a positive
    AUROC--gap slope (a ranking--axis \emph{trade-off}) and
    \textcolor[HTML]{16A085}{\textbf{green}} a negative slope
    (\emph{aligned}). Pearson's $r$ is reported in each panel
    (\textsuperscript{*}$p<0.05$,
    \textsuperscript{**}$p<0.01$,
    \textsuperscript{***}$p<0.001$).}
  \label{fig:reliability_facet_model}
\end{figure*}
\begin{tcolorbox}[colback=blue!5, colframe=blue!30, boxrule=0.5pt, left=4pt, right=4pt, top=3pt, bottom=3pt]
\small\textbf{Finding 2: Choosing a strong AUROC model does not necessarily sacrifice calibration, but it can sacrifice demographic equity.} 
\end{tcolorbox}

As shown in Fig. \ref{fig:reliability_facet_model}, not all deployment axes relate to AUROC in the same way. Calibration tends to improve with ranking performance, whereas rate-based demographic fairness gaps tend to worsen. For calibration, ECE is negatively correlated with AUROC, indicating that better ranking models are also better calibrated on average. This pattern holds especially strong within VM and VLM configurations.

Fairness behaves differently. At the family level, VM encoders achieve the highest mean AUROC. The reversal is clearest for age subgroup fairness: VM models have the largest age equalized-odds gap, while MLLMs have the smallest. Demographic parity, equal opportunity, and equalized odds gaps all show positive correlations with mean AUROC, with equal opportunity and equalized odds reaching statistical significance. Additionally, different adaptation methods can amplify the disparity. For the VLMs with both zero-shot and linear-probe results, linear probing improves AUROC for most models but increases the fairness gap for all six.

\begin{tcolorbox}[colback=blue!5, colframe=blue!30, boxrule=0.5pt, left=4pt, right=4pt, top=3pt, bottom=3pt]
\small\textbf{Finding 3: Glaucoma is the task cliff where retinal AI progress stalls.} 
\end{tcolorbox}
\begin{figure}
    \centering
    \includegraphics[width=1\linewidth]{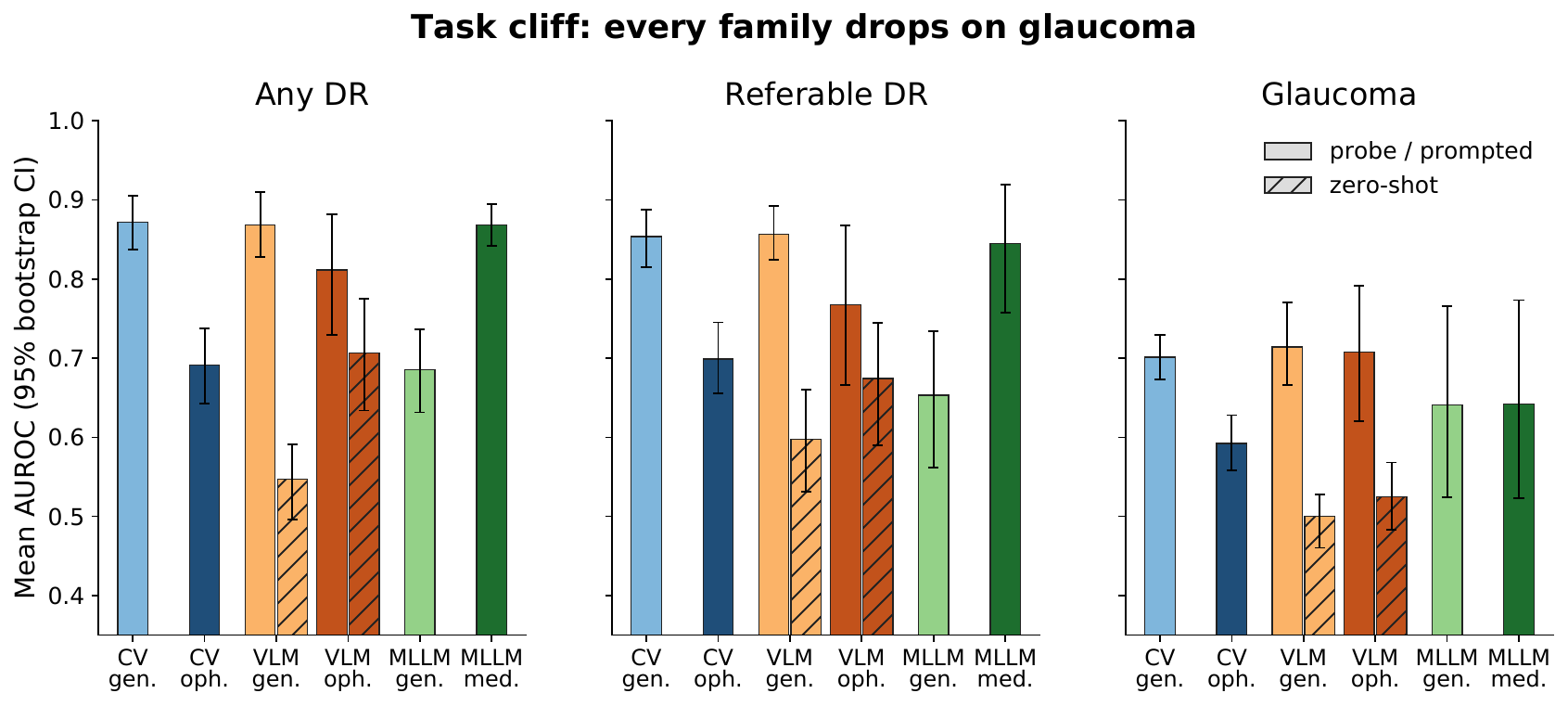}
    \caption{Glaucoma is the clearest task-level failure in FOCUS.}
    \label{fig:glaucoma_cliff}
\end{figure}

As shown in Fig.\ref{fig:glaucoma_cliff}, glaucoma is the obvious task domain where current retinal AI progress lags behind. Every adapted model family drops when moving from diabetic retinopathy to glaucoma. Generalist CV encoders fall from 0.867 AUROC on binary DR to 0.742 on glaucoma. Generalist VLMs with linear probing fall from 0.865 to 0.745. Ophthalmic CV and VLM encoders each lose at least 0.10 AUROC. The largest drop is for medical MLLMs, which match the strongest models on binary DR but fall from 0.868 to 0.656 on glaucoma.

This pattern shows that glaucoma is not solved by domain pretraining, medical instruction tuning, model scale, or lightweight adaptation. The best glaucoma group is generalist VLMs with a linear probe, reaching only 0.745 mean AUROC. On G1020, the hardest external glaucoma dataset, the best single configuration reaches 0.752 AUROC. Zero-shot VLMs fail most severely. Generalist VLM zero-shot evaluation reaches only near chance mean AUROC on glaucoma. Because glaucoma labels and imaging protocols vary substantially across datasets, this gap may reflect both intrinsic visual difficulty and dataset-level heterogeneity. Nevertheless, the consistency of the drop across model families indicates that current foundation-model strategies do not remove this bottleneck. This is clinically important because glaucoma is a leading cause of irreversible blindness \citep{tham2014global} and should not be overlooked in retinal foundation-model evaluation.

\begin{tcolorbox}[colback=blue!5, colframe=blue!30, boxrule=0.5pt, left=4pt, right=4pt, top=3pt, bottom=3pt]
\small\textbf{Finding 4: Specialization beats scale for MLLMs.} 
\end{tcolorbox}

\begin{figure}[t]
  \centering
  \includegraphics[width=1\linewidth]{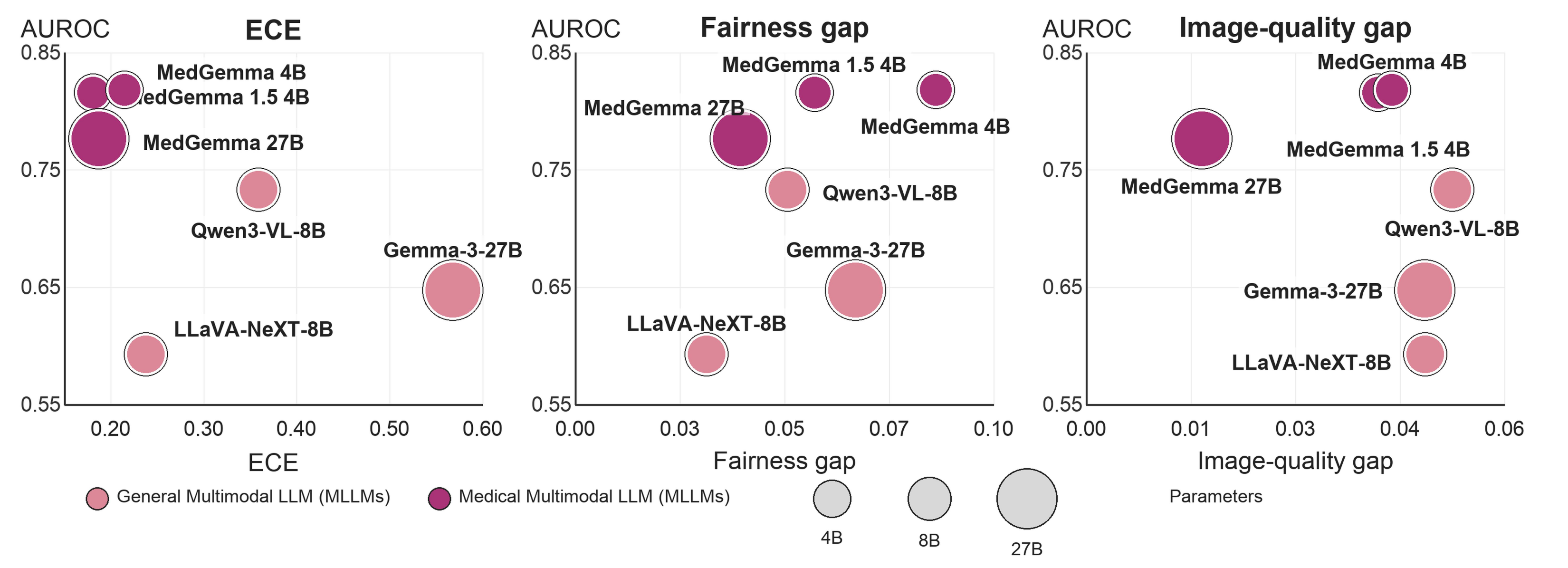}
  \caption{Medical specialization improves retinal MLLMs, but scaling from 4B to 27B does not guarantee better AUROC or uniformly better deployment behavior.}

  \label{fig:mllm-size}
\end{figure}

Figure~\ref{fig:mllm-size} compares two factors that are often conflated: parameter count and medical specialization. The result is clear: medical tuning gives the main gain, while additional scale does not reliably improve retinal performance. At fixed 27B scale, MedGemma-27B strongly outperforms Gemma-3-27B, increasing mean AUROC from 0.648 to 0.777 and reducing ECE from 0.57 to 0.19. This shows that medical specialization improves both ranking and calibration. Scaling within the medical family tells a different story. Within the evaluated MedGemma models, scaling from 4B to 27B does not improve mean AUROC. The gain is more consistent with domain specialization than with parameter count alone. 

The reliability axes also show that scale is not a clean solution. Fairness and image-quality gaps vary across models rather than following parameter count. The main takeaway is that medical specialization improves retinal MLLMs, but scaling from 4B to 27B does not guarantee better AUROC or uniformly better deployment behavior.

\begin{tcolorbox}[colback=blue!5, colframe=blue!30, boxrule=0.5pt, left=4pt, right=4pt, top=3pt, bottom=3pt]
\small\textbf{Finding 5: LoRA SFT improves calibration and thresholded decisions, not ranking.} 
\end{tcolorbox}

\begin{figure}[t]
  \centering
  \begin{minipage}[t]{0.49\linewidth}
    \centering
    \includegraphics[width=\linewidth]{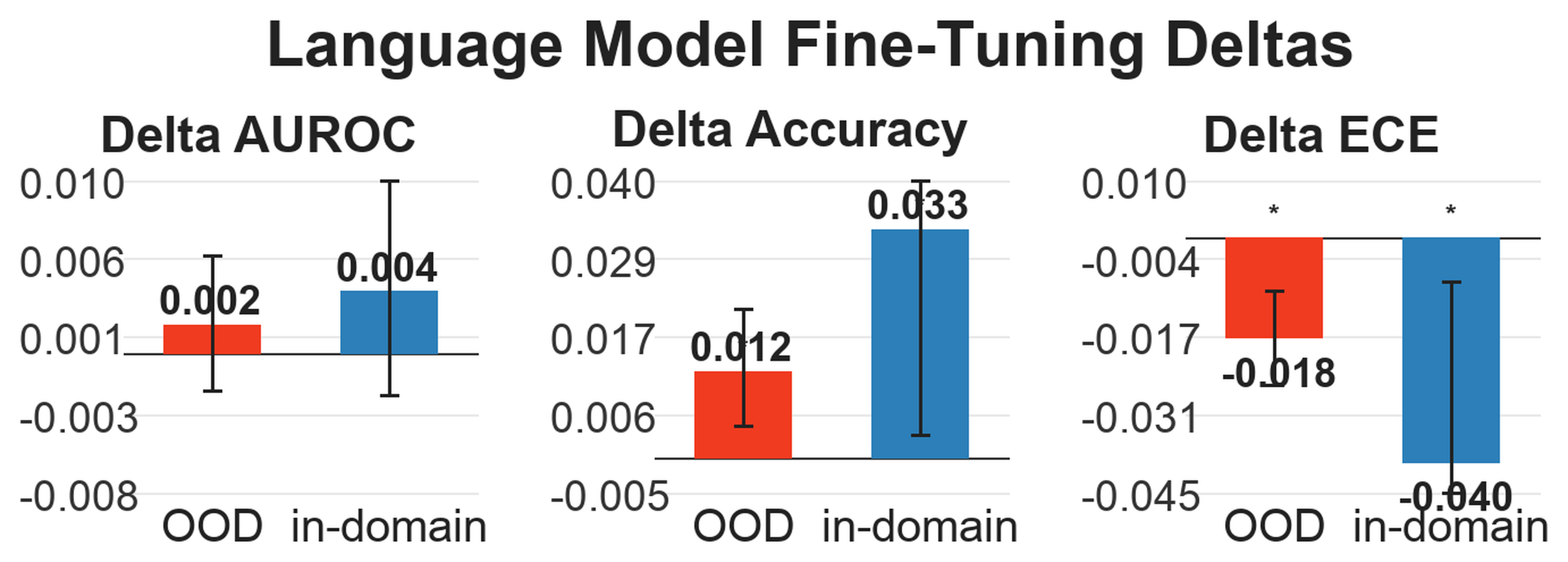}
  \end{minipage}\hfill
  \begin{minipage}[t]{0.5\linewidth}
    \centering
    \includegraphics[width=\linewidth]{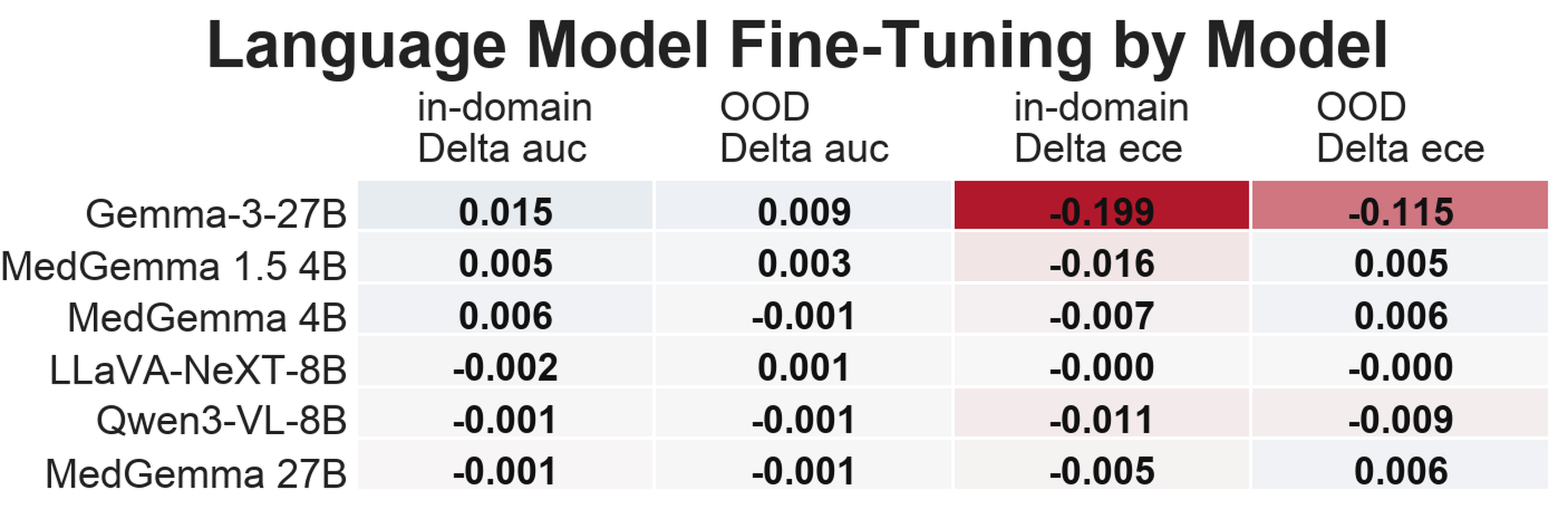}
  \end{minipage}
  \caption{\textbf{Language-model fine-tuning deltas.} Left: aggregate deltas versus base MLLMs on the same test dataset/task. Right: model-level deltas show that SFT effects are not uniform across base MLLMs or domains (negative $\Delta$ECE indicates improved calibration).}
  \label{fig:ft-delta-grid}
\end{figure}

As shown in Tab.~\ref{tab:ft-results} and Fig.~\ref{fig:ft-delta-grid} supervised fine-tuning helps MLLMs behave more consistently at the decision threshold, but it does not meaningfully improve average disease ranking. In-domain SFT improves accuracy by +0.033 and ECE by -0.040, both significant after correction. The same pattern mostly transfers OOD, with +0.012 accuracy and -0.018 ECE. In contrast, AUROC barely changes: +0.004 in-domain and +0.002 OOD, with non-significant q-values. This means SFT is not teaching the models a substantially better ordering of positive and negative retinal cases. Instead, it mainly changes how the model maps its existing signal into binary answers and confidence scores.

\begin{table}[t]
  \caption{Language-model fine-tuning transfer summary. Deltas are SFT metric minus matching base MLLM metric; negative $\Delta$ECE indicates improved calibration. Bootstrap \(q\)-values apply Benjamini--Hochberg correction within each metric family.}
  \label{tab:ft-results}
  \centering
  \scriptsize
  \resizebox{\linewidth}{!}{
  \begin{tabular}{lrrrrrrr}
    \toprule
    Test domain & AUROC & $\Delta$AUROC & \(q\) & $\Delta$Accuracy & \(q\) & $\Delta$ECE & \(q\) \\
    \midrule
    In-domain & 0.776 & 0.004 & 0.477 & 0.033 & 0.004 & -0.040 & 0.004 \\
    OOD & 0.724 & 0.002 & 0.521 & 0.012 & 0.005 & -0.018 & 0.004 \\
    \bottomrule
  \end{tabular}
  }
\end{table}

\paragraph{Robustness analyses.}
Across all 16 VM and encoder-VLM backbones, ten stratified bootstrap probe refits produced mean AUROC changes between $-0.012$ and $+0.004$ relative to the primary results. The two RETFound-DINOv2 variants showed larger accuracy changes despite relatively stable AUROC, indicating greater sensitivity of some thresholded comparisons (Table~\ref{tab:probe-stability}).

For MLLMs, adding an expert-role instruction or moving the output instruction before the clinical question changed AUROC by at most 0.020 and ECE by at most 0.055, while preserving the broad medical-versus-general trends (Table~\ref{tab:mllm-prompt-sensitivity}). Encoder-VLM zero-shot evaluation was more sensitive to label wording: MedSigLIP AUROC changed from 0.582 to 0.373 under the with/without formulation, whereas FLAIR improved from 0.720 to 0.822 under the finding-statement formulation. FLAIR and RET-CLIP remained the two strongest encoder VLMs under every tested formulation (Table~\ref{tab:vlm-prompt-sensitivity}).

The LoRA sensitivity analysis used joint three-task BRSET adapters for Gemma-3-27B and MedGemma-27B. In-domain AUROC and calibration improved across the tested settings, while OOD AUROC changes ranged from $-0.007$ to $+0.062$ for Gemma and from $+0.014$ to $+0.033$ for MedGemma (Table~\ref{tab:lora-sensitivity}). Full protocols and results are provided in Appendix~\ref{app:lora-sensitivity}.

\section{Discussion}

\benchmark{} is designed to test whether foundation-model adaptation to fundus photography generalizes beyond the narrow training setting. The current base benchmark shows that model interface and domain specialization interact in nontrivial ways: general VM encoders are strong and stable, medical MLLMs are competitive but not uniformly dominant, and VLMs can be weak in zero-shot form while becoming strong under linear probing. The language-model fine-tuning analysis adds a second layer: fine-tuning can improve in-domain thresholded behavior, but the same adapter may produce smaller or less reliable gains on external datasets.

These findings argue against a single retinal AI leaderboard. A clinically relevant benchmark should expose the trade-offs among ranking performance, calibration, subgroup behavior, image-quality robustness, and OOD transfer. \benchmark{} makes these axes explicit and keeps base and fine-tuned analyses separated so future work can measure whether adaptation helps, harms, or merely shifts failure modes.

\section{Limitations and Intended Use}

\benchmark{} is not a clinical validation study and should not be used to claim deployment readiness. Dataset labels are inherited from heterogeneous public or credentialed sources and may reflect different grading protocols. Some datasets lack demographic or image-quality metadata, limiting fairness and robustness analyses. Subgroup gaps are descriptive diagnostics, not causal fairness claims. Generative-model results depend on prompt templates, answer parsing, and token-probability availability. Benchmark scores should therefore be interpreted as evidence about controlled retrospective evaluation, not as evidence of safety for diagnosis or screening.

\section{Reproducibility}

Code and benchmark results are available through a \href{https://github.com/dsrestrepo/FOCUS-Retinal-Benchmark}{repository} and an interactive \href{https://huggingface.co/spaces/dsrestrepo/FOCUS-Retinal-Benchmark}{leaderboard dashboard}. The repository documentation describes the modular dataset and model interfaces, which can be used for dataset downloading/cleaning and to support additional fundus datasets, label mappings, tasks, and model families.

\begin{ack}
\textbf{Funding.}
This work was supported by the European Union's Horizon Europe research and innovation programme under the Marie Skłodowska-Curie COFUND grant agreement No.~101127936 (DeMythif.AI), and by France 2030 funding managed by the French National Research Agency (ANR) under the IA Cluster programme (DATAIA Cluster; grant No.~ANR-23-IACL-0003).

\textbf{Computational resources.}
This work was granted access to the HPC resources of the Jean Zay supercomputer operated by IDRIS (CNRS) and to the Ruche Mésocentre of Université Paris-Saclay.

\textbf{Competing interests.}
The authors declare no competing interests.
\end{ack}

\bibliographystyle{plainnat}
\bibliography{references}

\clearpage
\appendix
\section{Benchmark Card}

\paragraph{Name.}
\benchmark{}: \benchmarklong.

\paragraph{Purpose.}
\benchmark{} is intended for retrospective evaluation of retinal fundus models under dataset shift. The core question is whether foundation-model adaptation to fundus photography improves generalization and robustness, rather than only improving the test split of the dataset used for fine-tuning.

\paragraph{Intended use.}
Researchers may use \benchmark{} to compare model-family transfer, calibration, subgroup gaps, image-quality robustness, and supervised language-model fine-tuning behavior across harmonized retinal tasks.

\paragraph{Non-intended use.}
\benchmark{} is not intended for clinical deployment validation, diagnosis, screening triage, regulatory approval, or patient-level decision making.

\paragraph{Primary reporting axes.}
The benchmark reports AUROC, AUPRC, accuracy, F1, expected calibration error (ECE), fairness gaps for binary age/sex subgroups where metadata are available, and image-quality robustness gaps where quality strata are available. Metrics outside these implemented axes are omitted from the current report.

\section{Dataset and Task Matrix}
\label{apendix:datasets}

\begin{table}[H]
\centering
\small
\begin{tabular}{lccc}
\toprule
Dataset & Binary DR & Referable DR & Glaucoma \\
\midrule
BRSET & Yes & Yes & Yes \\
mBRSET & Yes & Yes & Yes \\
PAPILA & No & No & Yes \\
RFMiD & Yes & Yes & No \\
RFMiD 2.0 & Yes & Yes & No \\
IDRiD & Yes & Yes & No \\
Messidor-2 & Yes & Yes & No \\
REFUGE & No & No & Yes \\
G1020 & No & No & Yes \\
JSIEC1000 & Yes & Yes & No \\
\bottomrule
\end{tabular}
\caption{Configured dataset-task availability. A model is evaluated on a dataset-task pair only when the task label is available after harmonization.}
\label{tab:supp-dataset-task}
\end{table}

\begin{table}[H]
  \caption{Datasets configured in \benchmark. Tasks use the harmonized definitions in Sec.~\ref{sec:task-harmonization}; Fairness and robustness analyses are computed only when metadata are available.}
  \label{tab:datasets}
  \centering
  \setlength{\tabcolsep}{3pt}
  \begin{tabular}{llll}
    \toprule
    Dataset & Source & Tasks & Extra metadata \\
    \midrule
    BRSET & Brazil & Any DR, ref. DR, GON & Age, sex, image quality \\
    mBRSET & Brazil & Any DR, ref. DR, GON & Age, sex, image quality \\
    PAPILA & Spain & GON & Age, sex \\
    RFMiD & India & Any DR, ref. DR & -- \\
    RFMiD 2.0 & India & Any DR, ref. DR & -- \\
    IDRiD & India & Any DR, ref. DR & -- \\
    Messidor-2 & France & Any DR, ref. DR & -- \\
    REFUGE & China & GON & -- \\
    G1020 & Germany & GON & -- \\
    JSIEC1000 & China & Any DR, ref. DR & -- \\
    \bottomrule
  \end{tabular}
\end{table}

\subsection{Annotation provenance and operational mappings}
\label{app:label-provenance}
The implementation retains dataset-specific source annotations. The mappings in Table~\ref{tab:operational-labels} describe operational binary endpoints, not interchangeable clinical reference standards. In particular, a source-provided increased-cup/disc flag, a glaucoma/suspect diagnosis, and an expert glaucoma annotation need not denote the same underlying diagnostic criterion. 

\begin{table}[H]
\centering
\scriptsize
\begin{tabularx}{\linewidth}{lXXX}
\toprule
Dataset & Any-DR positive & Referable-DR positive & GON positive \\
\midrule
BRSET & \texttt{DR\_ICDR} grades 1--4 & Grade $\geq2$ or \texttt{macular\_edema}=1 & Source \texttt{increased\_cup\_disc} flag \\
mBRSET & \texttt{final\_icdr} grades 1--4 & Grade $\geq2$ or \texttt{final\_edema}=yes & Source \texttt{increased\_cdr} flag \\
PAPILA & -- & -- & Source diagnosis codes 1/2 (positive/suspect), versus code 0 \\
RFMiD & Source DR indicator $>0$ & Available grade $\geq2$ or positive source edema indicator & -- \\
RFMiD 2.0 & Source DR indicator $>0$ & Available grade $\geq2$ or positive source edema indicator & -- \\
IDRiD & Retinopathy grade $>0$ & Grade $\geq2$ or positive macular-edema-risk annotation & -- \\
Messidor-2 & Source/adjudicated DR grade $>0$ & Grade $\geq2$ or positive source DME annotation & -- \\
REFUGE & -- & -- & Source binary glaucoma annotation \\
G1020 & -- & -- & Source binary glaucoma annotation \\
JSIEC1000 & DR1/DR2/DR3 or suspected-PDR source folders & DR2/DR3 or suspected-PDR source folders & -- \\
\bottomrule
\end{tabularx}
\caption{Operational positive-label mappings in the released loaders/preparation rules. A dash denotes a task not configured for that dataset. Grade and edema fields are used only where available. These mappings do not remove annotation-protocol shift.}
\label{tab:operational-labels}
\end{table}

\section{Experiment Design}
\label{expdesign}

\subsection{Base benchmark}

The base benchmark evaluates three model families: VM encoders, VLMs, and MLLMs. The current result bundle contains 532 evaluated configurations: 190 VM configurations, 228 VLM configurations, and 114 MLLM configurations. Each configuration corresponds to one model, dataset, task, and evaluation method.

\paragraph{Data splits.}
For VM and VLM linear probing, the image encoder is frozen. The linear head is trained on the dataset's training split and evaluated on its test split. For VLM zero-shot and MLLM prompting, no task-specific fitting is performed; the model is evaluated on the test split. Missing labels are removed before metric calculation.

\paragraph{Task labels.}
All primary tasks are binary. Dataset-specific labels are harmonized into three target definitions: any diabetic retinopathy, referable diabetic retinopathy, and glaucoma.

\subsection{Fine-tuned MLLM transfer}

The fine-tuned analysis evaluates supervised LoRA adapters trained on BRSET or mBRSET for one task at a time and tests them on every configured dataset that supports the same task. The current SFT analysis contains 228 evaluations: 36 in-domain tests where the training and test datasets match, and 192 same-task out-of-domain (OOD) tests where they differ.

Each adaptation result is indexed by the training model, training dataset, training task, test dataset, test task, and evaluation split. The analysis keeps only same-task transfer pairs, because the current metrics are binary task metrics rather than cross-task semantic transfer metrics.

\section{Compute Resources}
\label{app:compute}

All experiments were run on a shared high-performance computing environment using slurm-style batch jobs. GPU jobs used NVIDIA H100-class accelerators with 80GB memory per GPU, CUDA 12.4, cuDNN 9.2, and NCCL 2.21. CPU-only jobs were used for dataset/model download orchestration and result aggregation. Table~\ref{tab:compute-resources} reports the resources requested by the released job scripts; wall-clock values are conservative job time limits rather than exact runtimes.

\begin{table}[H]
  \centering
  \small
  \setlength{\tabcolsep}{3pt}
  \begin{tabular}{lllll}
    \toprule
    Experiment stage & Worker type & Memory & Batch size & Time limit \\
    \midrule
    Dataset downloads & CPU, 4 cores & system memory & -- & 8h \\
    Model downloads & CPU, 4 cores & system memory & -- & 2h \\
    VM base evaluation & 1 H100 GPU, 24 CPU cores & 80GB GPU & 16 & 20h \\
    VLM base evaluation & 1 H100 GPU, 24 CPU cores & 80GB GPU & 16 & 20h \\
    MLLM base prompting & 1 H100 GPU, 24 CPU cores & 80GB GPU & 16 & 20h \\
    SFT LoRA training & 4 H100 GPUs, 48 CPU cores & 4$\times$80GB GPU & 1--4 & 18h \\
    SFT adapter evaluation & 1 H100 GPU, 24 CPU cores & 80GB GPU & 16 & 20h \\
    Base result analysis & CPU, 4 cores & system memory & -- & 2h \\
    FT result analysis/significance & CPU, 4 cores & system memory & -- & 2h \\
    \bottomrule
  \end{tabular}
  \caption{Compute resources used by the benchmark jobs. The SFT LoRA batch size is 1 with gradient accumulation 8 for 27B-class models and 4 with gradient accumulation 4 for smaller models. VM/VLM/MLLM base evaluation jobs iterate over the configured model--dataset--task--method grid for each family.}
  \label{tab:compute-resources}
\end{table}

\section{Inference Setup by Model Type}

\subsection{VM encoders}

For a frozen VM encoder \(f_\theta\), each image \(x_i\) is mapped to an embedding
\[
z_i=f_\theta(x_i).
\]
The benchmark trains a balanced logistic-regression probe on the training split:
\[
\ell_i = w^\top z_i + b,\qquad s_i=\sigma(\ell_i),
\]
where \(s_i\) is the positive-class score. The predicted label is the linear probe's class prediction \(\hat{y}_i\).

\paragraph{Linear probe configuration}

The same logistic-regression configuration is used for VM and VLM linear probing: L2 regularization with $C=1$, class-balanced weights, the \texttt{lbfgs} solver, and a maximum of 1,000 iterations. Extracted features are passed directly to the classifier without an additional feature-standardization stage.

The backbone remains frozen. VM feature extraction uses the CLS embedding; VLM probing uses the frozen image-tower embeddings. We fixed the probe configuration across models rather than selecting model-specific settings on validation data, to provide a common low-capacity evaluation protocol at the scale of the benchmark. Alternative regularization or preprocessing choices may change absolute scores and comparisons with small margins.

\subsection{VLMs}

VLMs are evaluated in two modes. In linear-probing mode, the image tower is frozen and the same logistic-regression formulation as above is applied to VLM image embeddings.

In zero-shot mode, the task provides a negative and positive text prompt. Let \(v_i\) be the image embedding and \(t_0,t_1\) the text embeddings. The implementation computes logits
\[
a_{ic}=100\, v_i^\top t_c,\qquad c\in\{0,1\},
\]
after the model-specific embedding normalization used by the VLM wrapper. The positive-class score is
\[
s_i=\frac{\exp(a_{i1})}{\exp(a_{i0})+\exp(a_{i1})},
\]
and \(\hat{y}_i=\arg\max_c a_{ic}\).

\paragraph{Prompt selection.}
The same fixed task-specific prompt pair is used across encoder VLMs and applicable datasets.

\begin{description}

\item[Any DR, negative:]
A fundus photograph of a healthy retina without diabetic retinopathy.
\item[Any DR, positive:]
A fundus photograph of a retina with diabetic retinopathy.
\item[Referable DR, negative:]
A fundus photograph showing no referable diabetic retinopathy.
\item[Referable DR, positive:]
A fundus photograph showing referable diabetic retinopathy.
\item[GON, negative:]
A fundus photograph of a healthy eye without glaucoma.
\item[GON, positive:]
A fundus photograph showing signs of glaucoma.
\end{description}

Sensitivity to alternative prompt formulations is reported in Appendix~\ref{app:vlm-prompt-sensitivity}.

\subsection{MLLMs}
\label{llm:inference}

MLLMs are evaluated with task-specific prompts that require a binary answer. For an image \(x_i\) and prompt \(q_i\), the model defines a next-token distribution over the vocabulary after the chat template has been applied:
\[
p_\theta(v \mid x_i,q_i)
=
\operatorname{softmax}(z_i)_v,
\]
where \(z_i\) are the logits for the first generated token. We extract the probabilities assigned to affirmative and negative answer tokens from this first-token distribution. To reduce tokenizer-format dependence, each answer word is evaluated using single-token variants with no prefix, a leading space, and a leading newline:
\[
\mathcal{Y}=\{\texttt{yes},\texttt{ Yes},\texttt{\textbackslash nYes},
\texttt{Yes},\texttt{ yes},\texttt{\textbackslash nyes}\},
\]
\[
\mathcal{N}=\{\texttt{no},\texttt{ No},\texttt{\textbackslash nNo},
\texttt{No},\texttt{ no},\texttt{\textbackslash nno}\}.
\]
Only variants that tokenize to a single token are used. The affirmative and negative first-token probabilities are
\[
p_i^{+}=\max_{v\in\mathcal{Y}} p_\theta(v\mid x_i,q_i),
\qquad
p_i^{-}=\max_{v\in\mathcal{N}} p_\theta(v\mid x_i,q_i).
\]
The positive-class score used for AUROC, AUPRC, and calibration is
\[
s_i=p_i^{+}.
\]
The hard predicted label is parsed from the generated text by taking the first standalone yes/no answer:
\[
\hat{y}_i =
\begin{cases}
1 & \text{if the parsed answer is yes},\\
0 & \text{if the parsed answer is no}.
\end{cases}
\]
Rows without a valid parsed yes/no label or without a valid first-token affirmative score are excluded from metric calculation.

\paragraph{MLLM prompts.}
For binary diabetic retinopathy:
\begin{quote}
Based on the fundus image, does this eye show diabetic retinopathy?

Respond with \textbf{yes} if any level of diabetic retinopathy is present, or \textbf{no} otherwise. Respond only with ``yes'' or ``no'' without additional commentary.
\end{quote}

For referable diabetic retinopathy:
\begin{quote}
Based on the fundus image, does this eye show referable diabetic retinopathy?

Respond with \textbf{yes} if the image should be referred for diabetic retinopathy evaluation or treatment, or \textbf{no} otherwise. Respond only with ``yes'' or ``no'' without additional commentary.
\end{quote}

For glaucoma:
\begin{quote}
Based on the fundus image, does this eye show glaucoma or glaucomatous optic disc changes?

Respond with \textbf{yes} if glaucoma or suspicious glaucomatous changes are present, or \textbf{no} otherwise. Respond only with ``yes'' or ``no'' without additional commentary.
\end{quote}

\subsection{Fine-tuned MLLMs}

Fine-tuned MLLMs use the same inference pipeline as base MLLMs, with an SFT LoRA adapter loaded into the base model before generation. The same parser and scoring pipeline is used for base and fine-tuned models so that deltas reflect adaptation rather than evaluation changes.

\section{Language Model Fine-tuning Details}
\label{llm:ft}

\paragraph{Training grid.}
The current FT grid trains the six configured MLLMs on BRSET and mBRSET for any diabetic retinopathy, referable diabetic retinopathy, and glaucoma. The evaluated MLLMs are Gemma-3-27B, MedGemma-4B, MedGemma-1.5-4B, LLaVA-NeXT-8B, Qwen3-VL-8B, and MedGemma-27B.

\paragraph{Adapter configuration.}
SFT uses LoRA adapters targeting all linear modules with rank \(r=16\), no bias terms, and causal language modeling as the PEFT task type. The implementation uses LoRA dropout 0.05, bf16 training, gradient checkpointing, and the configured 16-bit quantization mode.

\paragraph{Objective.}
Supervised fine-tuning optimizes the assistant answer tokens with cross-entropy:
\[
\mathcal{L}_{\mathrm{SFT}}
=-\sum_{j\in\mathcal{A}_i}\log p_\theta(a_{ij}\mid x_i,q_i,a_{i,<j}),
\]
where \(x_i\) is the fundus image, \(q_i\) is the task prompt, and \(\mathcal{A}_i\) indexes the assistant answer tokens. The assistant target is the binary answer text. The SFT configuration uses one epoch, learning rate \(2\times 10^{-5}\), LoRA alpha 16, batch size 1 with gradient accumulation 8 for 27B-class models, and batch size 4 with gradient accumulation 4 for smaller models.

\section{Metric Definitions}
\label{app:metrics}

Let \(D=\{(y_i,\hat{y}_i,s_i)\}_{i=1}^n\), where \(y_i\in\{0,1\}\) is the ground-truth label, \(\hat{y}_i\in\{0,1\}\) is the predicted label, and \(s_i\in[0,1]\) is the positive-class score. Metrics are set to missing when the required labels, predictions, scores, or class diversity are unavailable.

\paragraph{Accuracy.}
\[
\operatorname{Accuracy}=\frac{1}{n}\sum_{i=1}^n \mathbb{1}[\hat{y}_i=y_i].
\]

\paragraph{Precision, recall, and F1.}
With \(TP\), \(FP\), and \(FN\) computed for the positive class,
\[
\operatorname{Precision}=\frac{TP}{TP+FP},\qquad
\operatorname{Recall}=\frac{TP}{TP+FN},
\]
\[
\operatorname{F1}=\frac{2\,\operatorname{Precision}\,\operatorname{Recall}}
{\operatorname{Precision}+\operatorname{Recall}}.
\]

\paragraph{AUROC.}
Let \(P=\{i:y_i=1\}\) and \(N=\{j:y_j=0\}\). AUROC is the probability that a positive example receives a higher score than a negative example, with ties receiving half credit:
\[
\operatorname{AUROC}=
\frac{1}{|P||N|}
\sum_{i\in P}\sum_{j\in N}
\left(\mathbb{1}[s_i>s_j]+\frac{1}{2}\mathbb{1}[s_i=s_j]\right).
\]

\paragraph{AUPRC.}
The implementation uses scikit-learn average precision. Sorting examples by descending \(s_i\), let \(P_k\) and \(R_k\) be precision and recall after the \(k\)-th threshold. Then
\[
\operatorname{AUPRC}=\sum_k (R_k-R_{k-1})P_k.
\]

\paragraph{Expected calibration error.}
The implemented ECE is positive-class probability calibration, not correctness-confidence calibration. Divide \([0,1]\) into \(M=10\) equal-width bins \(B_m\) according to \(s_i\). For each non-empty bin,
\[
\operatorname{score}(B_m)=\frac{1}{|B_m|}\sum_{i\in B_m}s_i,\qquad
\operatorname{pos}(B_m)=\frac{1}{|B_m|}\sum_{i\in B_m}y_i.
\]
ECE is
\[
\operatorname{ECE}=\sum_{m=1}^M \frac{|B_m|}{n}
\left|\operatorname{score}(B_m)-\operatorname{pos}(B_m)\right|.
\]
Lower ECE is better.

\section{Fairness Metrics}
\label{meteics:fairness}

For a binary subgroup attribute \(a_i\in\{0,1\}\), let \(G_g=\{i:a_i=g\}\). Gaps are reported as max-minus-min across the two groups, so lower is better. These are descriptive subgroup diagnostics and should not be interpreted as causal fairness estimates.

\paragraph{Demographic parity gap.}
\[
\Delta_{\mathrm{DP}} =
\left|
\Pr(\hat{Y}=1\mid G_1)-\Pr(\hat{Y}=1\mid G_0)
\right|.
\]

\paragraph{Accuracy gap.}
\[
\Delta_{\mathrm{Acc}}=
\left|
\operatorname{Accuracy}(G_1)-\operatorname{Accuracy}(G_0)
\right|.
\]

\paragraph{Equal opportunity gap.}
With \(\operatorname{TPR}(G)=\Pr(\hat{Y}=1\mid Y=1,G)\),
\[
\Delta_{\mathrm{TPR}}=
\left|
\operatorname{TPR}(G_1)-\operatorname{TPR}(G_0)
\right|.
\]

\paragraph{False-positive-rate gap.}
With \(\operatorname{FPR}(G)=\Pr(\hat{Y}=1\mid Y=0,G)\),
\[
\Delta_{\mathrm{FPR}}=
\left|
\operatorname{FPR}(G_1)-\operatorname{FPR}(G_0)
\right|.
\]

\paragraph{Equalized odds gap.}
The implemented equalized odds summary is the maximum of the true-positive-rate and false-positive-rate gaps:
\[
\Delta_{\mathrm{EO}}=\max(\Delta_{\mathrm{TPR}},\Delta_{\mathrm{FPR}}).
\]
It is not the average of TPR and FPR gaps.

\paragraph{AUROC gap.}
\[
\Delta_{\mathrm{AUROC}}=
\left|
\operatorname{AUROC}(G_1)-\operatorname{AUROC}(G_0)
\right|,
\]
when both subgroup AUROCs are defined.

\section{Subgroup Construction}

\paragraph{Age.}
Age is binarized within the analyzed result rows using the median valid age as the threshold:
\[
a_i=\mathbb{1}[\operatorname{age}_i>\operatorname{median}(\operatorname{age})].
\]
This avoids selecting a very young cutoff that would produce tiny minority groups in older retinal cohorts.

\paragraph{Sex.}
Sex is normalized to male/female using the dataset-specific metadata column when available. Rows with unrecognized or missing sex values are excluded from sex subgroup calculations.

\paragraph{Image quality.}
Robustness uses binary image-quality strata. For BRSET, the provided image-quality label is binarized into adequate versus inadequate quality. For mBRSET, the provided final quality flag is binarized into accepted versus not accepted quality.

\paragraph{Minimum group sizes.}
The default minimum subgroup size is 10. Image-quality robustness uses the configured minimum of 5. If fewer than two groups remain or the smaller group is below the threshold, the metric is skipped in the summary tables.

\section{Robustness Metrics}

Robustness is computed overall for each model, dataset, and task, rather than inside every fairness subgroup. This avoids creating many small partitions. For a binary image-quality attribute \(Q\in\{0,1\}\) and a metric \(M\),
\[
\Delta_{Q,M}=\left|M(Q=1)-M(Q=0)\right|.
\]
The implementation reports image-quality gaps for accuracy, AUROC, AUPRC, F1, and ECE when both quality groups are available.

\section{Fine-tuning Delta Metrics}

For a fine-tuned adapter \(A\), test dataset \(d\), and task \(t\), the FT analysis joins the adapter metric with the base MLLM metric for the same model, dataset, and task. For any metric \(M\),
\[
\Delta M(A,d,t)=M_{\mathrm{FT}}(A,d,t)-M_{\mathrm{base}}(\operatorname{model}(A),d,t).
\]
Positive deltas are better for AUROC, AUPRC, accuracy, and F1. Negative deltas are better for ECE. The analysis also records
\[
\operatorname{OOD}(A,d)=\mathbb{1}[\operatorname{train}(A)\neq d].
\]

\subsection{Uncertainty for fine-tuning deltas}

The FT significance analysis uses paired predictions rather than only aggregate summaries. For each adapter--test comparison \(c\), examples are aligned by image identifier between the fine-tuned adapter and the corresponding base MLLM:
\[
\mathcal{D}_c=\{(y_i,\hat{y}^{\mathrm{base}}_i,s^{\mathrm{base}}_i,\hat{y}^{\mathrm{FT}}_i,s^{\mathrm{FT}}_i)\}_{i=1}^{n_c},
\]
where \(s_i\) is the positive-class score. The observed paired delta is
\[
\widehat{\delta}_{c,M}
=M(\{y_i,\hat{y}^{\mathrm{FT}}_i,s^{\mathrm{FT}}_i\}_{i=1}^{n_c})
-M(\{y_i,\hat{y}^{\mathrm{base}}_i,s^{\mathrm{base}}_i\}_{i=1}^{n_c}).
\]
For per-comparison uncertainty, bootstrap sample \(b\) draws \(n_c\) indices with replacement from \(\{1,\ldots,n_c\}\) and recomputes \(\widehat{\delta}^{(b)}_{c,M}\). The 95\% interval is the empirical \([2.5,97.5]\) percentile interval. The two-sided bootstrap p-value is
\[
p_{\mathrm{boot}}
=2\min\left(
\frac{1+\sum_b\mathbb{1}[\widehat{\delta}^{(b)}_{c,M}\le 0]}{B+1},
\frac{1+\sum_b\mathbb{1}[\widehat{\delta}^{(b)}_{c,M}\ge 0]}{B+1}
\right),
\]
clipped at 1. For the domain-level summaries in the main paper, the analysis bootstraps the mean of the paired deltas across SFT adapter--test comparisons within each test domain and metric. Reported \(q\)-values apply Benjamini--Hochberg correction within each metric family.

\section{Sensitivity and Stability Analyses}
\label{app:sensitivity}

\subsection{Linear-probe refitting stability}
\label{app:probe-stability}

We expanded the stability analysis to all 16 VM and encoder-VLM backbones across the ten benchmark datasets, using applicable dataset--task pairs. Each frozen-feature probe was refitted on ten class-stratified bootstrap samples of the training set, with a different random-state seed for each refit, and evaluated on the unchanged test split.

For each refit, metric changes relative to the corresponding primary result were macro-averaged across applicable dataset--task evaluations. Table~\ref{tab:probe-stability} reports the mean and standard deviation of these run-level changes. Fairness summaries average available equalized-odds, accuracy, and AUROC gaps; quality summaries average available image-quality metric gaps. These summaries use only evaluations with the required metadata.

\begin{table}[H]
\centering
\scriptsize
\setlength{\tabcolsep}{3pt}
\resizebox{\linewidth}{!}{%
\begin{tabular}{lccccc}
\toprule
Model & $\Delta$AUROC & $\Delta$Accuracy & $\Delta$ECE & $\Delta$Fairness gap & $\Delta$Quality gap \\
\midrule
VMs \\
\midrule
ViT-L/16
& $-0.004\pm0.005$ & $+0.002\pm0.003$ & $-0.005\pm0.003$ & $-0.003\pm0.006$ & $+0.009\pm0.013$ \\
DINOv2
& $-0.007\pm0.004$ & $-0.000\pm0.007$ & $-0.005\pm0.002$ & $-0.001\pm0.007$ & $+0.001\pm0.008$ \\
DINOv3
& $-0.006\pm0.006$ & $-0.000\pm0.003$ & $-0.004\pm0.002$ & $+0.000\pm0.003$ & $+0.007\pm0.010$ \\
RETFound-CFP
& $-0.008\pm0.003$ & $+0.000\pm0.005$ & $-0.004\pm0.004$ & $+0.003\pm0.006$ & $-0.003\pm0.013$ \\
RETFound-OCT
& $-0.005\pm0.005$ & $-0.003\pm0.007$ & $-0.003\pm0.006$ & $+0.013\pm0.007$ & $+0.010\pm0.012$ \\
RETFound-MEH
& $-0.002\pm0.005$ & $-0.002\pm0.006$ & $-0.001\pm0.003$ & $+0.005\pm0.007$ & $+0.003\pm0.016$ \\
RETFound-Shanghai
& $-0.005\pm0.004$ & $+0.000\pm0.004$ & $-0.002\pm0.003$ & $-0.005\pm0.007$ & $-0.001\pm0.034$ \\
RETFound-DINOv2-MEH
& $-0.003\pm0.009$ & $+0.195\pm0.162$ & $+0.000\pm0.001$ & $+0.003\pm0.011$ & $-0.003\pm0.014$ \\
RETFound-DINOv2-Shanghai
& $-0.003\pm0.009$ & $+0.061\pm0.110$ & $+0.000\pm0.000$ & $+0.001\pm0.003$ & $-0.002\pm0.009$ \\
VisionFM
& $-0.012\pm0.006$ & $-0.001\pm0.001$ & $-0.010\pm0.003$ & $+0.004\pm0.005$ & $+0.015\pm0.021$ \\
\midrule
VLMs \\
\midrule
MedSigLIP 
& $-0.007\pm0.006$ & $+0.002\pm0.003$ & $-0.004\pm0.003$ & $-0.001\pm0.004$ & $+0.004\pm0.006$ \\
SigLIP2
& $-0.005\pm0.005$ & $-0.003\pm0.004$ & $-0.002\pm0.003$ & $+0.002\pm0.004$ & $+0.000\pm0.010$ \\
CLIP
& $-0.004\pm0.003$ & $+0.001\pm0.005$ & $-0.004\pm0.003$ & $-0.000\pm0.006$ & $+0.005\pm0.014$ \\
EyeCLIP
& $+0.004\pm0.010$ & $+0.018\pm0.019$ & $-0.000\pm0.002$ & $-0.005\pm0.010$ & $+0.005\pm0.016$ \\
RET-CLIP
& $-0.002\pm0.003$ & $-0.001\pm0.003$ & $-0.004\pm0.003$ & $+0.002\pm0.002$ & $+0.007\pm0.005$ \\
FLAIR
& $-0.003\pm0.002$ & $+0.000\pm0.002$ & $-0.001\pm0.004$ & $+0.001\pm0.002$ & $+0.009\pm0.007$ \\
\bottomrule
\end{tabular}%
}
\caption{Linear-probe refitting stability. Entries are mean changes $\pm$ standard deviation across ten bootstrap refits, relative to the primary results. Positive AUROC/accuracy changes and negative ECE/gap changes indicate improvement.}
\label{tab:probe-stability}
\end{table}

Mean AUROC changes range from $-0.012$ to $+0.004$. Accuracy is less stable for the two RETFound-DINOv2 variants despite their small AUROC changes. These results support the stability of broad ranking-based comparisons under the evaluated perturbations, while indicating that some thresholded comparisons require caution.

\subsection{Zero-shot encoder-VLM prompt sensitivity}
\label{app:vlm-prompt-sensitivity}

The primary zero-shot evaluation uses fixed task-specific label descriptions without model-specific prompt optimization. We evaluated three alternative prompt templates across all six encoder VLMs:

\begin{itemize}
    \item \textbf{Label only} (``No [condition]'' / ``[condition]''),
    \item \textbf{With/without} (``A retina without [condition]'' / ``A retina with [condition]''), and
    \item \textbf{Finding statement} (``A retinal image showing no signs of [condition]'' / ``A retinal image showing signs of [condition]'').
\end{itemize}

The same formulation was applied across models.

\begin{table}[H]
\centering
\small
\setlength{\tabcolsep}{4pt}
\begin{tabular}{lcccc}
\toprule
Model & Base & Label only & With/without & Finding statement \\
\midrule
MedSigLIP & 0.582 & 0.466 & 0.373 & 0.406 \\
SigLIP2   & 0.547 & 0.519 & 0.494 & 0.548 \\
CLIP      & 0.515 & 0.563 & 0.547 & 0.567 \\
EyeCLIP   & 0.511 & 0.530 & 0.521 & 0.504 \\
RET-CLIP  & 0.689 & 0.618 & 0.604 & 0.572 \\
FLAIR     & 0.720 & 0.696 & 0.798 & 0.822 \\
\bottomrule
\end{tabular}
\caption{Zero-shot encoder-VLM prompt sensitivity. Values are AUROC, macro-averaged across applicable dataset--task evaluations. Base denotes the primary prompt formulation.}
\label{tab:vlm-prompt-sensitivity}
\end{table}

Sensitivity is model-dependent. MedSigLIP decreases from 0.582 to 0.373 AUROC under the with/without formulation, while FLAIR increases from 0.720 to 0.822 under the finding-statement formulation. FLAIR and RET-CLIP remain the two strongest models under every tested formulation.

\subsection{MLLM prompt sensitivity}
\label{app:mllm-prompt-sensitivity}

The primary prompts were manually designed by two experts, including an ophthalmologist, to request a clear binary response. We evaluated two variants: adding an expert-role instruction (``You are an expert ophthalmologist'') and moving the yes/no output instruction before the clinical question. The clinical question and requested binary decision were otherwise unchanged.

\begin{table}[H]
\centering
\scriptsize
\setlength{\tabcolsep}{6pt}
\begin{tabular}{llcccc}
\toprule
Model & Prompt & AUROC & ECE & Fairness gap & Quality gap \\
\midrule
Gemma-3-27B & Base & 0.651 & 0.556 & 0.072 & 0.047 \\
 & Expert role & 0.651 & 0.545 & 0.062 & 0.039 \\
 & Instruction first & 0.638 & 0.520 & 0.068 & 0.058 \\
\midrule
MedGemma-4B & Base & 0.806 & 0.205 & 0.070 & 0.043 \\
 & Expert role & 0.804 & 0.191 & 0.067 & 0.037 \\
 & Instruction first & 0.786 & 0.200 & 0.069 & 0.045 \\
\midrule
MedGemma-1.5-4B & Base & 0.806 & 0.192 & 0.054 & 0.047 \\
 & Expert role & 0.790 & 0.197 & 0.060 & 0.056 \\
 & Instruction first & 0.797 & 0.200 & 0.064 & 0.042 \\
\midrule
LLaVA-NeXT-8B & Base & 0.593 & 0.672 & 0.050 & 0.057 \\
 & Expert role & 0.573 & 0.676 & 0.041 & 0.056 \\
 & Instruction first & 0.609 & 0.665 & 0.044 & 0.055 \\
\midrule
Qwen3-VL-8B & Base & 0.722 & 0.373 & 0.057 & 0.054 \\
 & Expert role & 0.722 & 0.318 & 0.056 & 0.054 \\
 & Instruction first & 0.721 & 0.342 & 0.052 & 0.041 \\
\midrule
MedGemma-27B & Base & 0.770 & 0.180 & 0.039 & 0.017 \\
 & Expert role & 0.769 & 0.177 & 0.049 & 0.019 \\
 & Instruction first & 0.767 & 0.187 & 0.047 & 0.022 \\
\bottomrule
\end{tabular}
\caption{MLLM prompt sensitivity. Metrics are macro-averaged across applicable dataset--task evaluations. Fairness and quality use the metadata-backed composite gaps defined in Appendix~\ref{app:probe-stability}. Higher AUROC and lower ECE/gaps are preferred.}
\label{tab:mllm-prompt-sensitivity}
\end{table}

The maximum absolute changes were 0.020 AUROC, 0.055 ECE, 0.011 fairness gap, and 0.013 quality gap. The broad medical-versus-general trends were preserved, although close comparisons and reliability metrics varied with prompt wording.

\subsection{MLLM response-parsing audit}
\label{app:mllm-parsing-audit}

Of 58,332 audited responses, 42 (0.072\%) could not be mapped to yes/no by the strict response parser (regex matching "yes" or "no" tokens). For five of the six models, the parsed generated answer agreed with the binary-normalized first-token yes/no argmax probabilities in 98.4--99.9\% of cases.

LLaVA-NeXT was the exception: it often generated whitespace before the answer. A valid decoded yes/no answer could therefore occur after the first generated position, while the primary score was still extracted from the first-position distribution. Its probability-based results are consequently sensitive to token-position and answer-format conventions.

\subsection{LoRA configuration sensitivity}
\label{app:lora-sensitivity}

We evaluated Gemma-3-27B and MedGemma-27B using one joint BRSET adapter per configuration, trained across all three tasks. These supplementary experiments differ from the task-specific adapters used in the primary fine-tuning analysis. We varied rank, alpha, and target modules around the reference configuration ($r=16$, $\alpha=16$, all linear modules), changing one factor at a time.

In-domain results average the three BRSET tasks. OOD results average the external dataset--task evaluations shared by all configurations. Deltas are relative to the corresponding unadapted backbone.

\begin{table}[H]
\centering
\scriptsize
\setlength{\tabcolsep}{5pt}
\begin{tabular}{llrrrr}
\toprule
Model & Configuration & $\Delta$ID AUROC & $\Delta$ID ECE & $\Delta$OOD AUROC & $\Delta$OOD ECE \\
\midrule
Gemma-3-27B & Reference & $+0.089$ & $-0.109$ & $+0.021$ & $-0.265$ \\
 & Rank 8 & $+0.081$ & $-0.106$ & $+0.021$ & $-0.287$ \\
 & Rank 32 & $+0.092$ & $-0.112$ & $+0.026$ & $-0.283$ \\
 & Alpha 8 & $+0.052$ & $-0.087$ & $+0.007$ & $-0.172$ \\
 & Alpha 32 & $+0.096$ & $-0.110$ & $+0.062$ & $-0.326$ \\
 & Query/value & $+0.057$ & $-0.092$ & $-0.007$ & $-0.149$ \\
 & All attention & $+0.057$ & $-0.088$ & $-0.000$ & $-0.187$ \\
\midrule
MedGemma-27B & Reference & $+0.021$ & $-0.030$ & $+0.033$ & $-0.006$ \\
 & Rank 8 & $+0.021$ & $-0.029$ & $+0.019$ & $-0.006$ \\
 & Rank 32 & $+0.021$ & $-0.030$ & $+0.023$ & $-0.006$ \\
 & Alpha 8 & $+0.019$ & $-0.030$ & $+0.024$ & $-0.014$ \\
 & Alpha 32 & $+0.025$ & $-0.034$ & $+0.028$ & $-0.012$ \\
 & Query/value & $+0.008$ & $-0.015$ & $+0.017$ & $+0.000$ \\
 & All attention & $+0.014$ & $-0.025$ & $+0.014$ & $-0.011$ \\
\bottomrule
\end{tabular}
\caption{LoRA configuration sensitivity for joint three-task BRSET adapters. Reference uses rank 16, alpha 16, and all linear target modules. Positive AUROC deltas and negative ECE deltas indicate improvement.}
\label{tab:lora-sensitivity}
\end{table}

Rank had relatively small effects, while alpha and target modules changed the magnitude of gains, particularly for Gemma. In-domain AUROC and calibration improved across the tested settings, whereas OOD ranking effects were more heterogeneous. These results support the qualitative conclusion that calibration benefits are more consistent than cross-dataset ranking improvements within this supplementary setting.

\section{Summary Tables}

\subsection{AUROC by task and model type}

\begin{table}[H]
\centering
\small
\begin{tabular}{lrrr}
\toprule
Model type & Binary DR & Glaucoma & Referable DR \\
\midrule
General Vision Encoder Model (VM) & 0.867 & 0.742 & 0.850 \\
Ophthalmic Vision Encoder Model (VM) & 0.690 & 0.591 & 0.697 \\
General Multimodal LLM (MLLM) & 0.685 & 0.630 & 0.651 \\
Medical Multimodal LLM (MLLM) & 0.868 & 0.656 & 0.844 \\
General Dual Encoder Vision Language Model (VLM-encoders) & 0.706 & 0.616 & 0.724 \\
Ophthalmic Dual Encoder Vision Language Model (VLM-encoders) & 0.756 & 0.630 & 0.721 \\
\bottomrule
\end{tabular}
\caption{Mean AUROC by task and model type in the current benchmark result bundle.}
\label{tab:supp-task-model-type}
\end{table}

\subsection{Language model fine-tuning summary}

\begin{table}[H]
\centering
\small
\begin{tabular}{lrrrrr}
\toprule
Domain & AUROC & AUPRC & Accuracy & F1 & ECE \\
\midrule
In-domain & 0.776 & 0.619 & 0.826 & 0.511 & 0.164 \\
OOD & 0.724 & 0.513 & 0.675 & 0.402 & 0.290 \\
\bottomrule
\end{tabular}
\caption{Mean SFT metrics by test domain. The corresponding delta estimates are summarized in the main paper.}
\label{tab:supp-ft-domain}
\end{table}

\begin{table}[H]
\centering
\scriptsize
\begin{tabular}{llrrrr}
\toprule
Domain & Metric & Mean delta & 95\% CI low & 95\% CI high & \(q\) \\
\midrule
In-domain & AUROC & 0.004 & -0.002 & 0.010 & 0.477 \\
In-domain & Accuracy & 0.033 & 0.003 & 0.076 & 0.004 \\
In-domain & ECE & -0.040 & -0.085 & -0.008 & 0.004 \\
OOD & AUROC & 0.002 & -0.002 & 0.006 & 0.521 \\
OOD & Accuracy & 0.012 & 0.005 & 0.021 & 0.005 \\
OOD & ECE & -0.018 & -0.026 & -0.009 & 0.004 \\
\bottomrule
\end{tabular}
\caption{Bootstrap uncertainty for domain-level SFT deltas. Positive ECE deltas indicate worse calibration, while negative ECE deltas indicate improved calibration. Intervals are estimated by paired resampling over matched base and fine-tuned predictions.}
\label{tab:supp-ft-significance}
\end{table}

\subsection{Dataset and model coverage}

\begin{figure}[H]
  \centering
  \includegraphics[width=\linewidth]{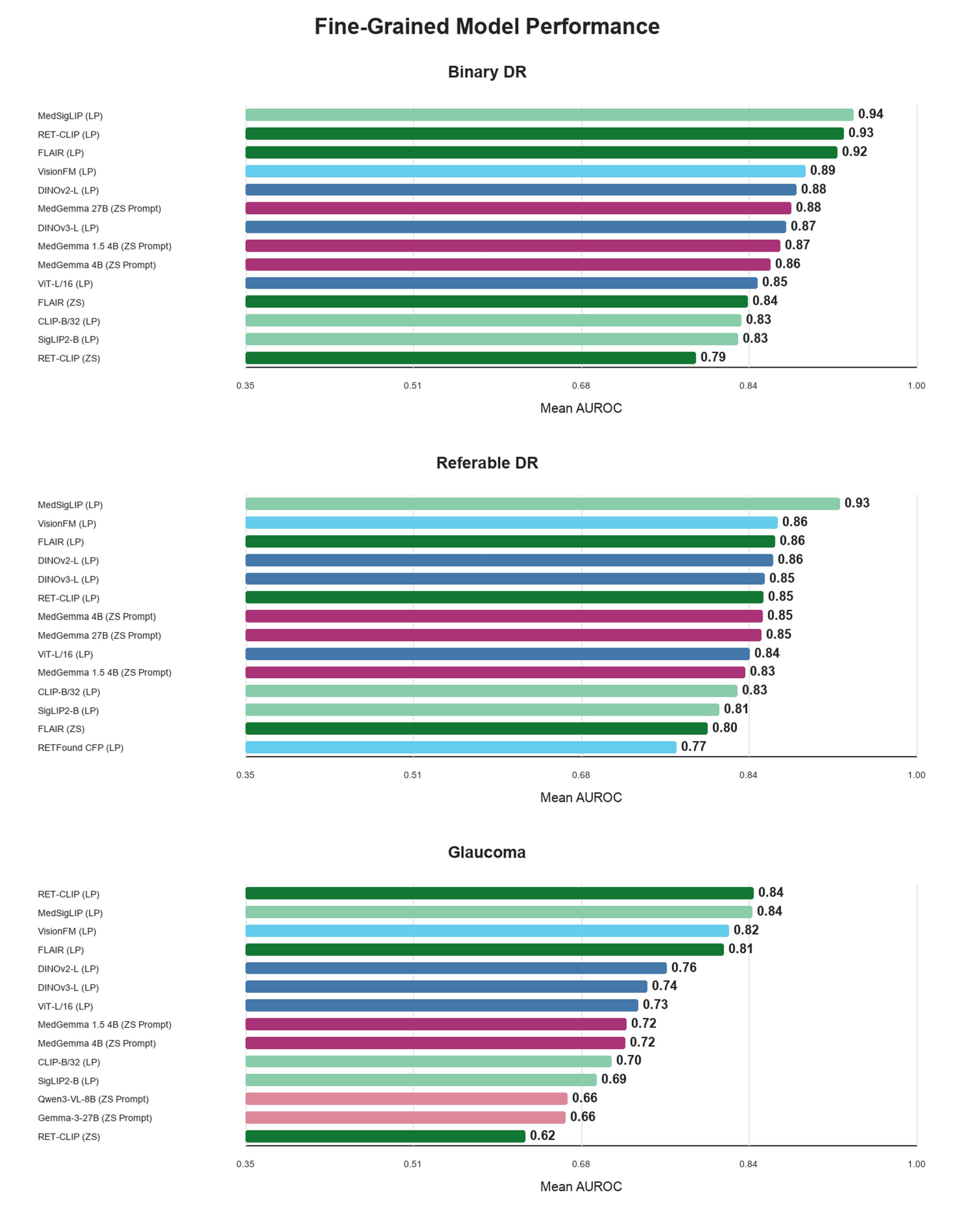}
  \caption{\textbf{Fine-grained model performance.} Task-specific model rankings expose differences that are hidden by family averages. This larger version is kept in the supplement because it is useful for inspection but too dense for the main narrative.}
  \label{fig:supp-fine-grained-performance}
\end{figure}

\begin{figure}[H]
  \centering
  \includegraphics[width=\linewidth]{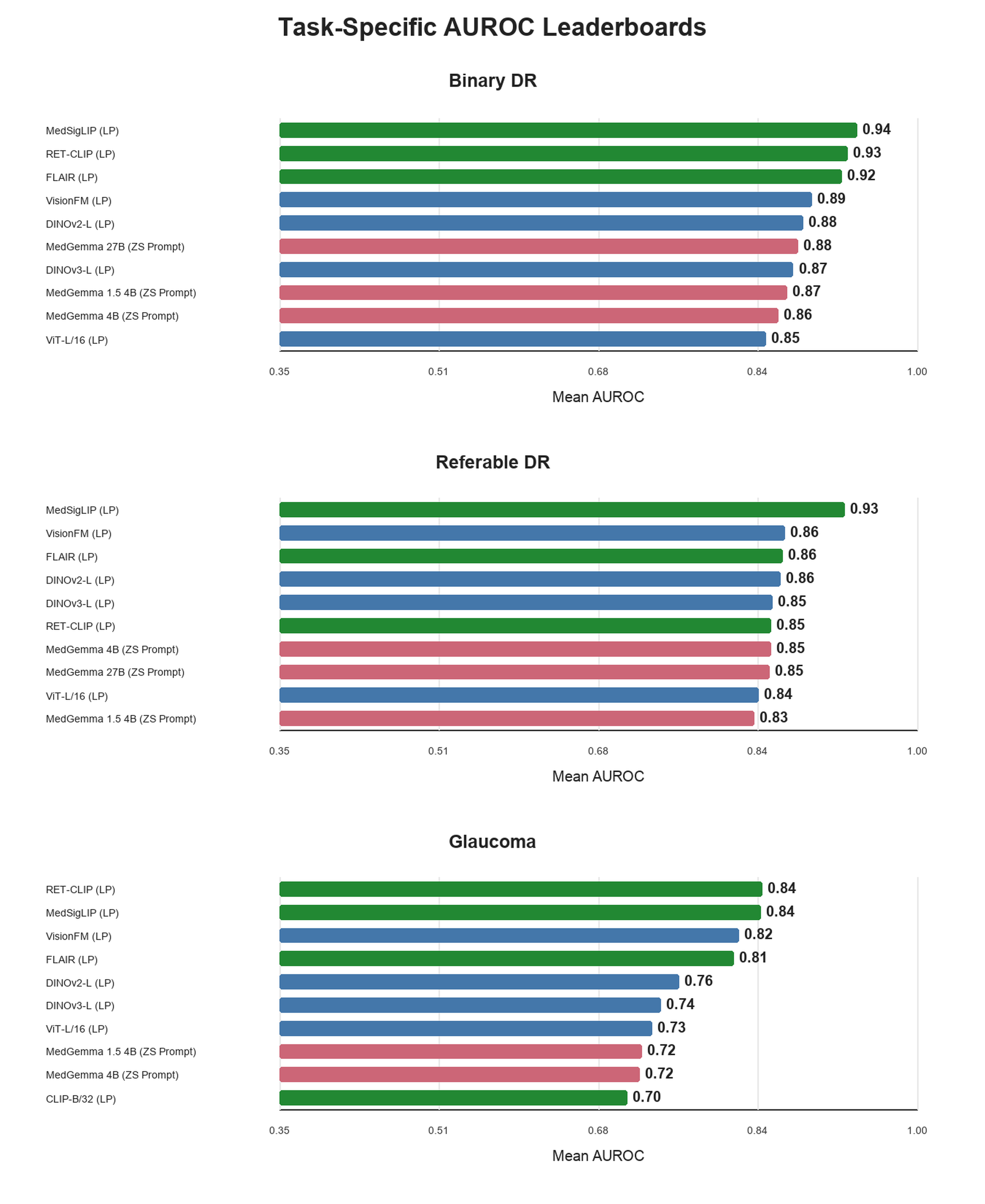}
  \caption{\textbf{Task-specific leaderboards.} Top AUROC configurations differ by task, reinforcing that a benchmark for fundus foundation models should expose disease-specific strengths rather than collapse all results into a single score.}
  \label{fig:supp-task-leaderboards}
\end{figure}

\begin{figure}[H]
  \centering
  \includegraphics[width=\linewidth]{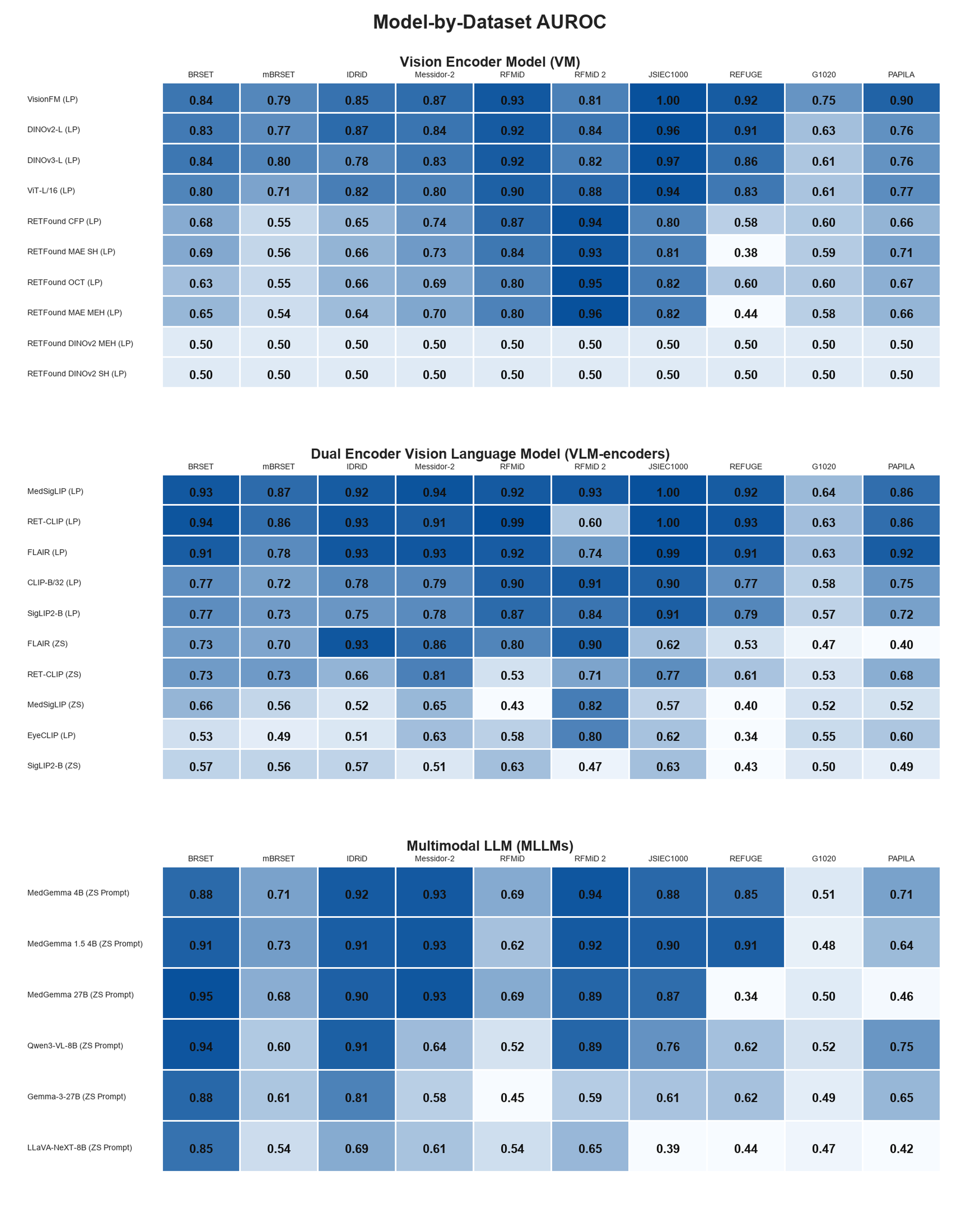}
  \caption{\textbf{Model-by-dataset AUROC heatmap.} Rows are model/method configurations and columns are datasets. This plot is intended to reveal dataset-specific failures that are hidden by averaging over all datasets.}
  \label{fig:supp-model-dataset}
\end{figure}

\begin{figure}[H]
  \centering
  \includegraphics[width=\linewidth]{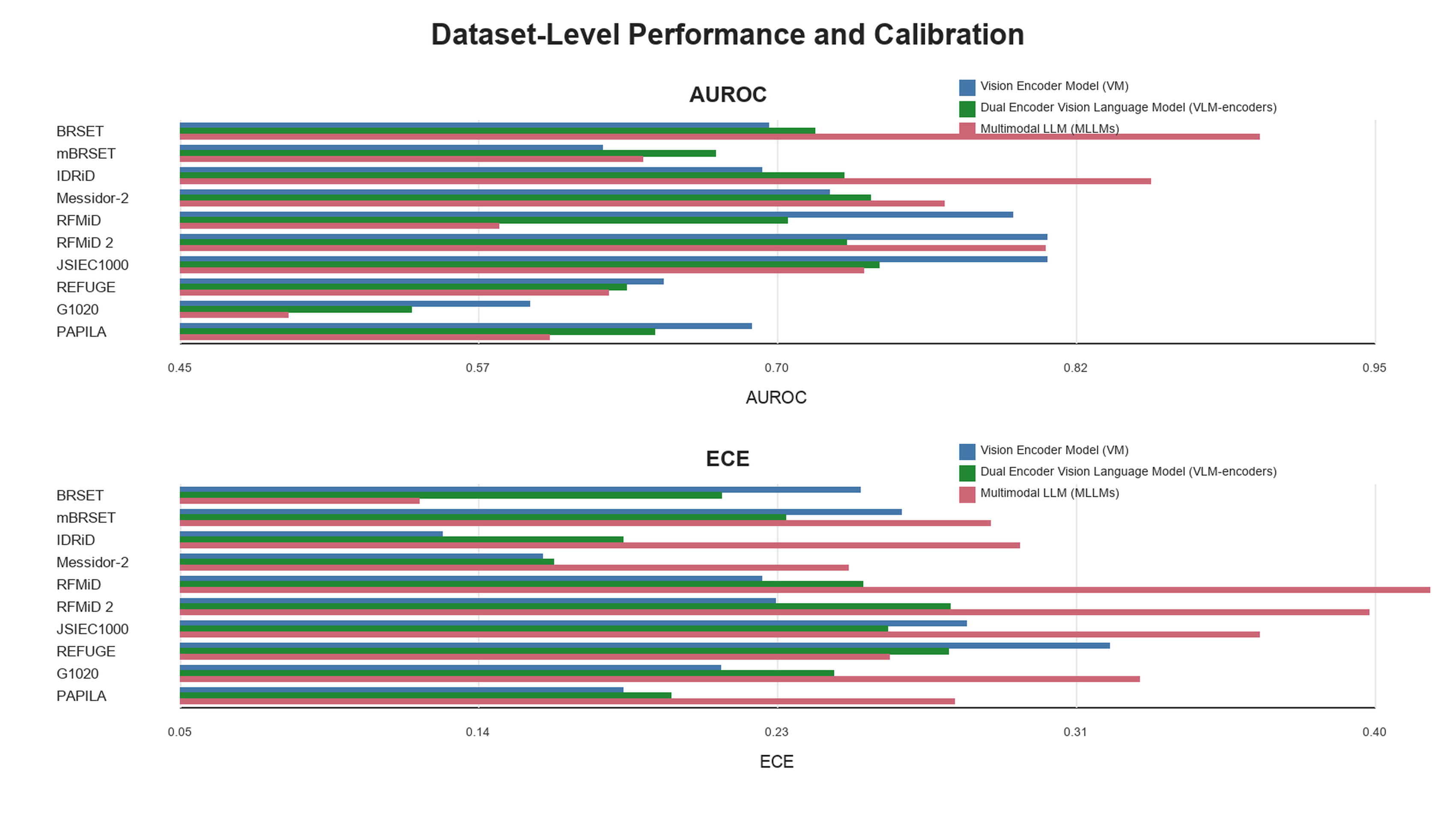}
  \caption{\textbf{Dataset-level family metrics.} AUROC and ECE are shown by dataset and model family.}
  \label{fig:supp-dataset-family}
\end{figure}

\subsection{Calibration diagnostics}

\begin{figure}[H]
  \centering
  \includegraphics[width=\linewidth]{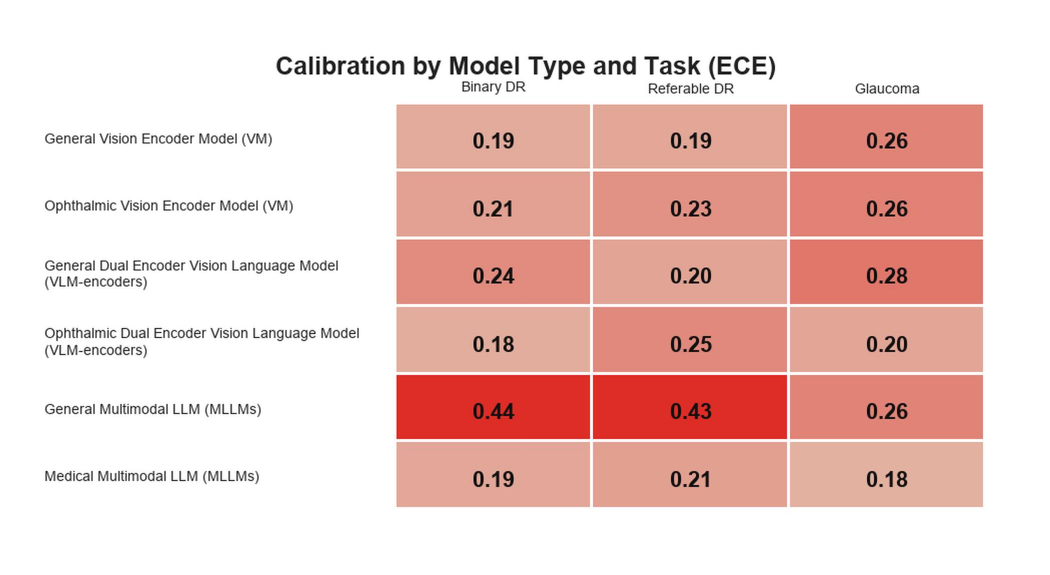}
  \caption{\textbf{Calibration by model type.} Mean ECE is summarized by dataset and task for each model type. Empty cells indicate unavailable dataset-task combinations or missing valid scores.}
  \label{fig:supp-calibration-model-type}
\end{figure}

\begin{figure}[H]
  \centering
  \includegraphics[width=\linewidth]{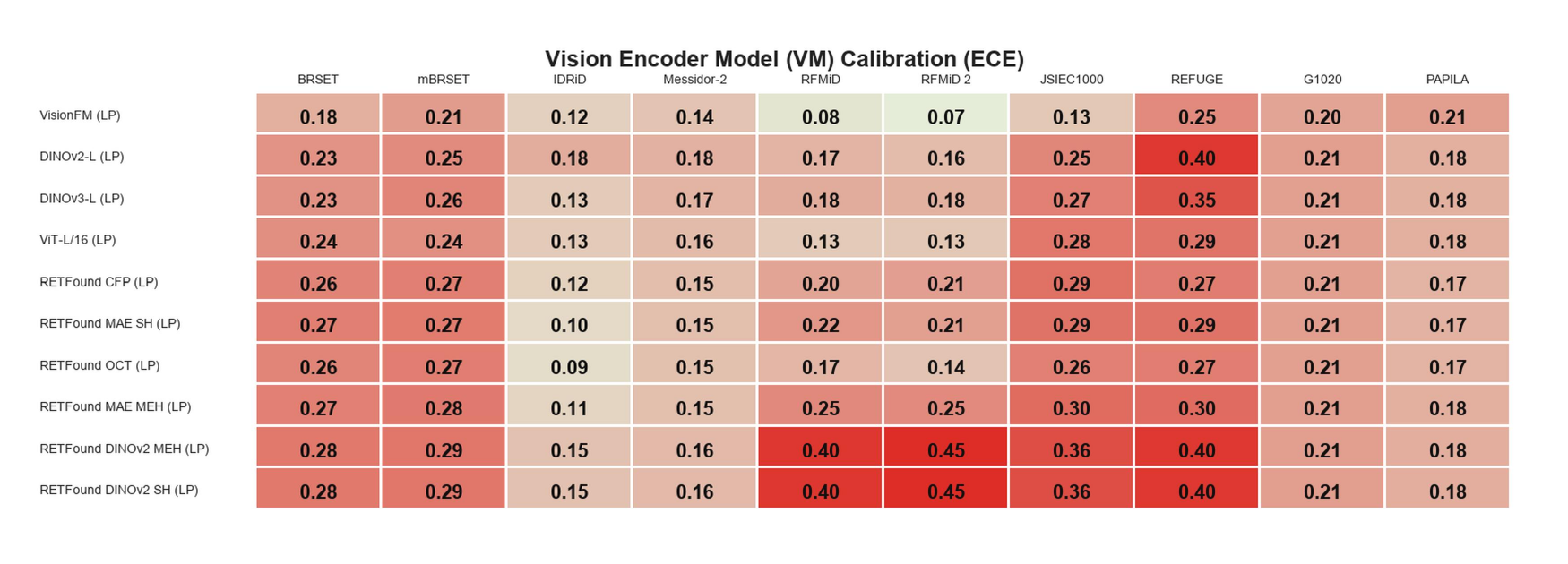}
  \caption{\textbf{VM calibration grid.} Mean ECE for VM encoder configurations, organized by dataset and task.}
  \label{fig:supp-calibration-cv}
\end{figure}

\begin{figure}[H]
  \centering
  \includegraphics[width=\linewidth]{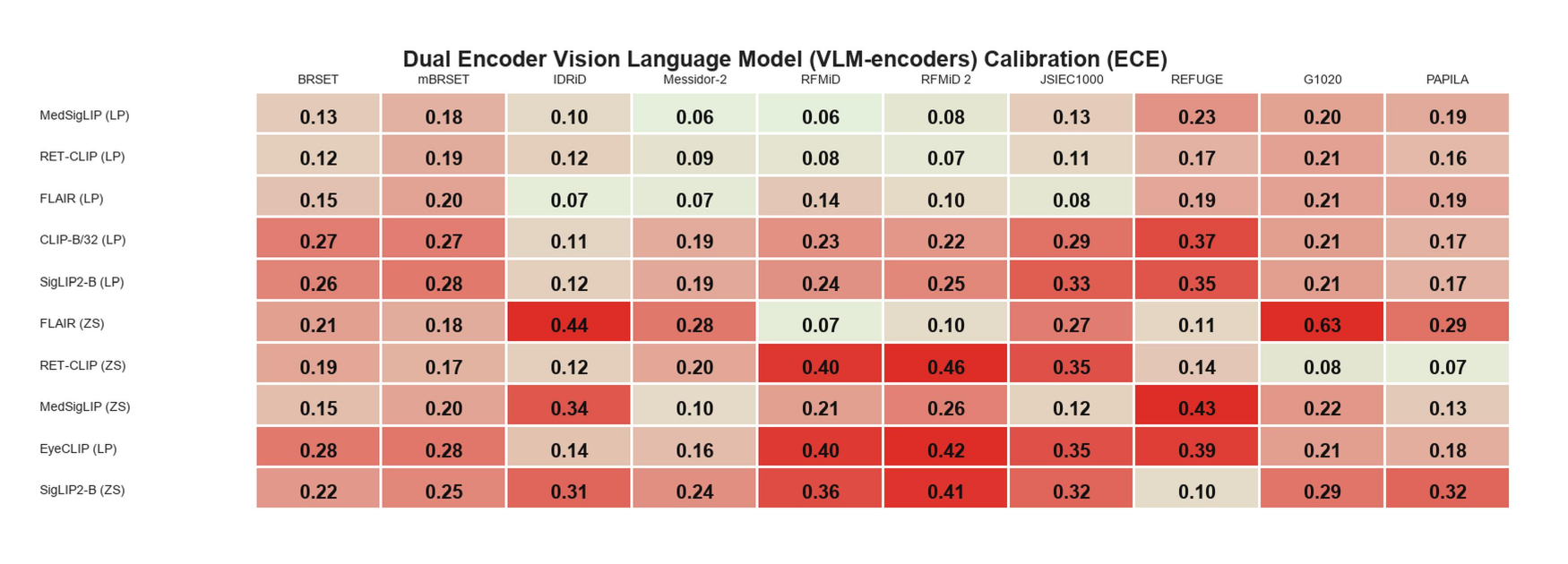}
  \caption{\textbf{VLM calibration grid.} Mean ECE for VLM configurations, including both zero-shot and linear-probing results where available.}
  \label{fig:supp-calibration-vlm}
\end{figure}

\begin{figure}[H]
  \centering
  \includegraphics[width=\linewidth]{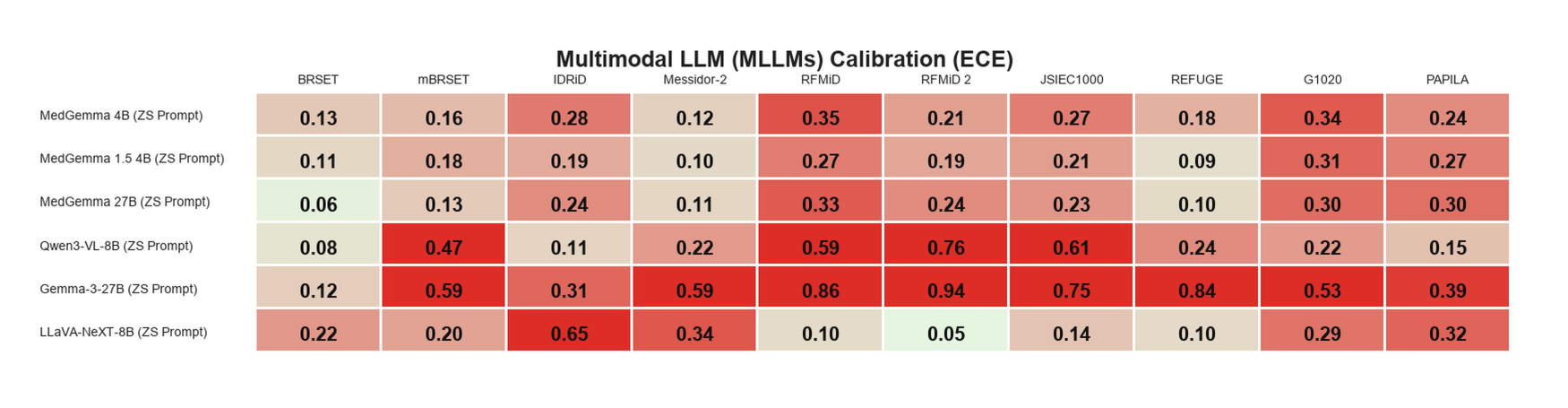}
  \caption{\textbf{MLLM calibration grid.} Mean ECE for prompted MLLM configurations by dataset and task.}
  \label{fig:supp-calibration-mllm}
\end{figure}

\subsection{Calibration curves}

The following sheets restore the raw calibration-curve diagnostics. Each sheet fixes one model type and lays out dataset/task panels; lines show the top configurations within that model type. These figures are intentionally supplementary because they are large and are meant for error analysis rather than quick comparison.

\begin{figure}[H]
  \centering
  \includegraphics[width=\linewidth]{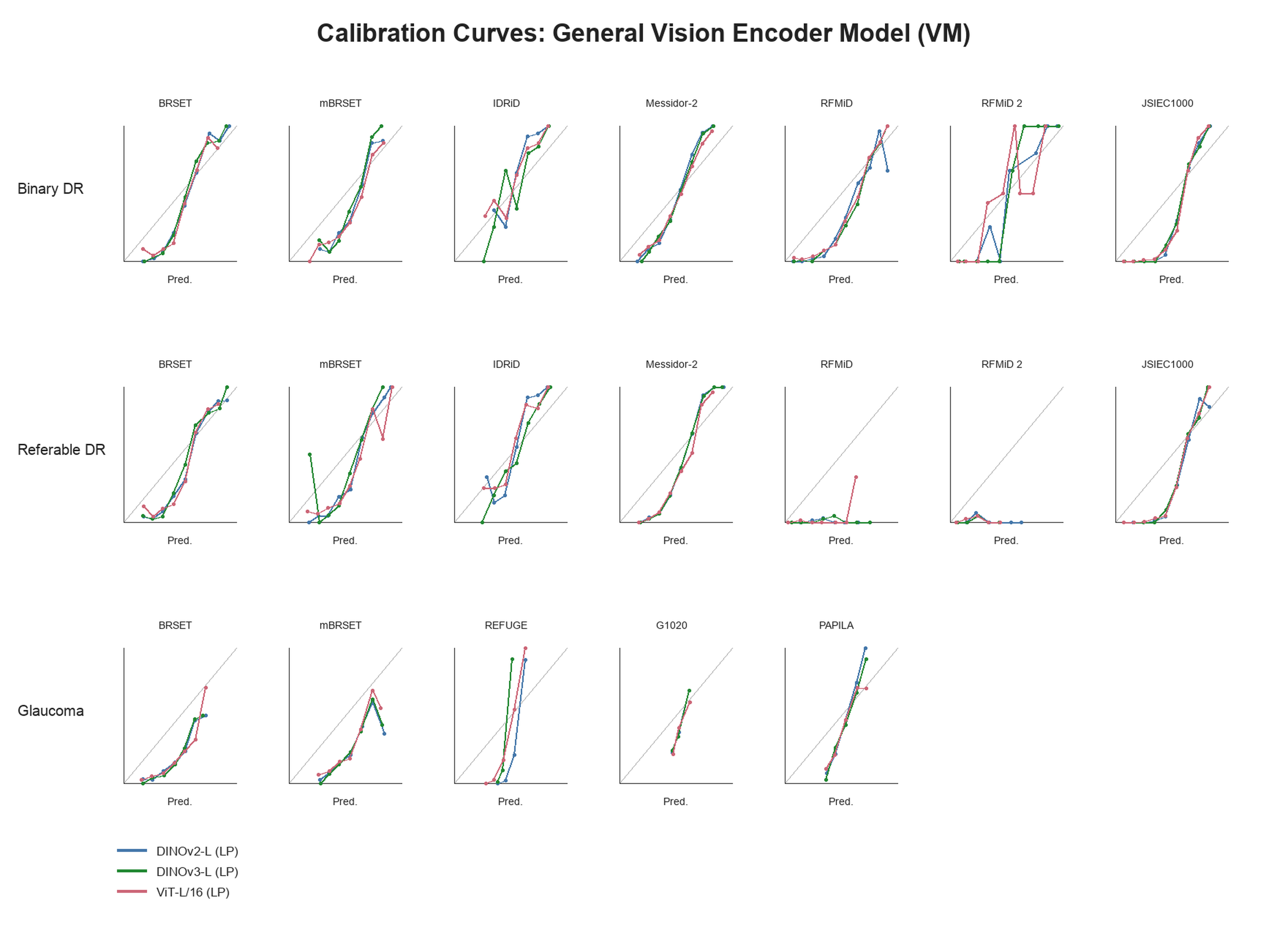}
  \caption{\textbf{Calibration curves for general VM encoders.} Each panel is a dataset/task pair; the diagonal indicates perfect calibration.}
  \label{fig:supp-calibration-curves-cv-general}
\end{figure}

\begin{figure}[H]
  \centering
  \includegraphics[width=\linewidth]{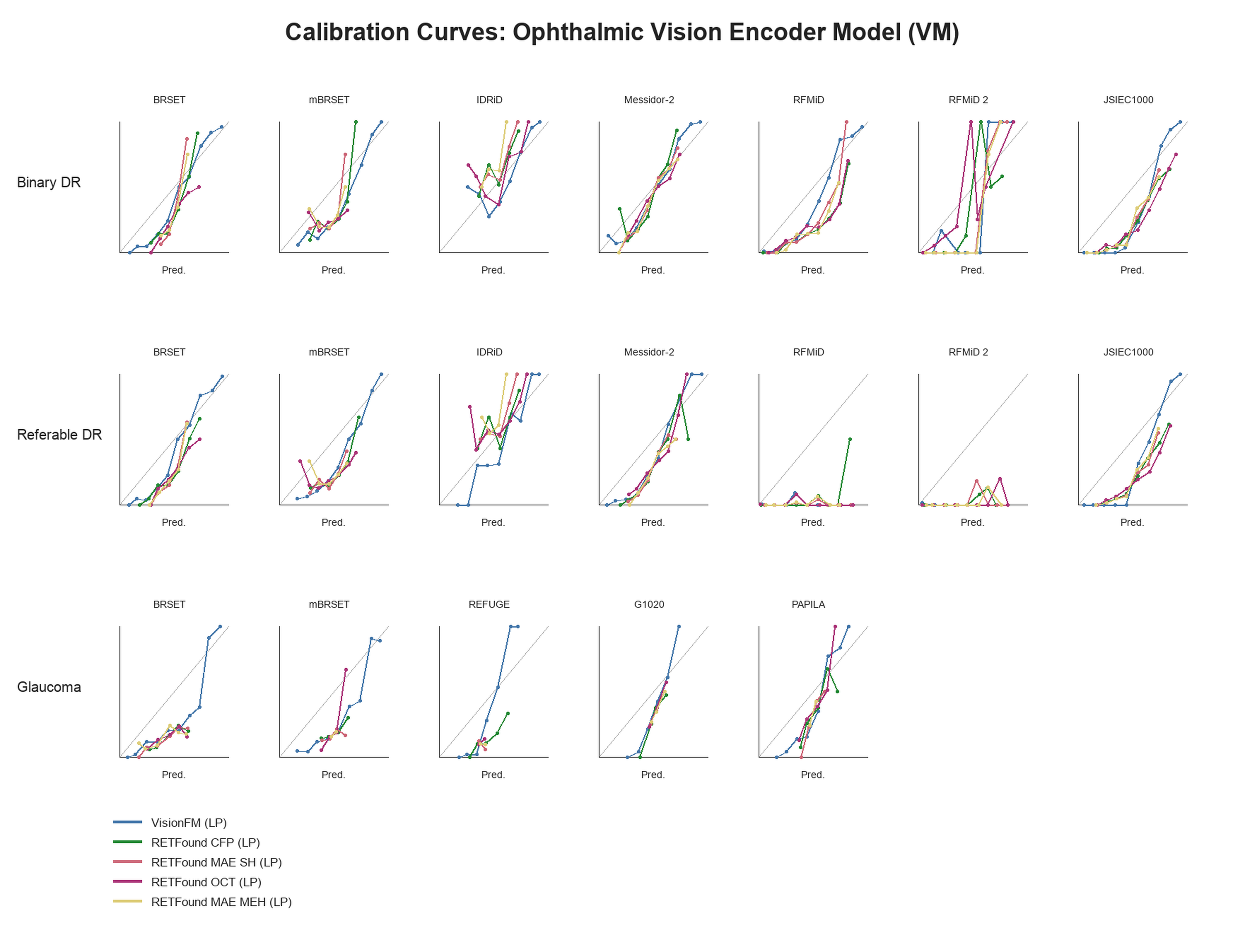}
  \caption{\textbf{Calibration curves for ophthalmic VM encoders.} Each panel is a dataset/task pair; the diagonal indicates perfect calibration.}
  \label{fig:supp-calibration-curves-cv-ophthalmo}
\end{figure}

\begin{figure}[H]
  \centering
  \includegraphics[width=\linewidth]{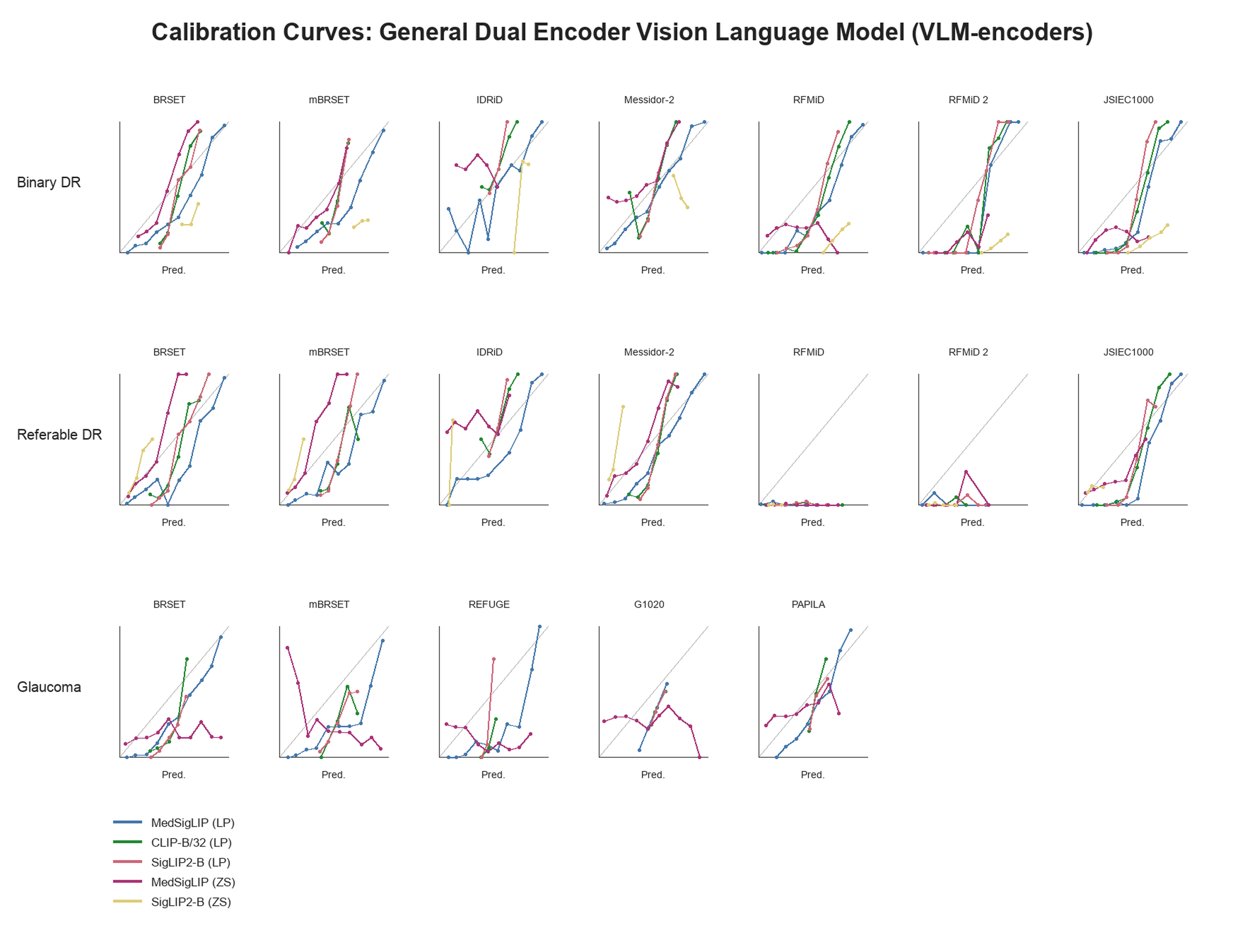}
  \caption{\textbf{Calibration curves for general VLMs.} Each panel is a dataset/task pair; the diagonal indicates perfect calibration.}
  \label{fig:supp-calibration-curves-vlm-general}
\end{figure}

\begin{figure}[H]
  \centering
  \includegraphics[width=\linewidth]{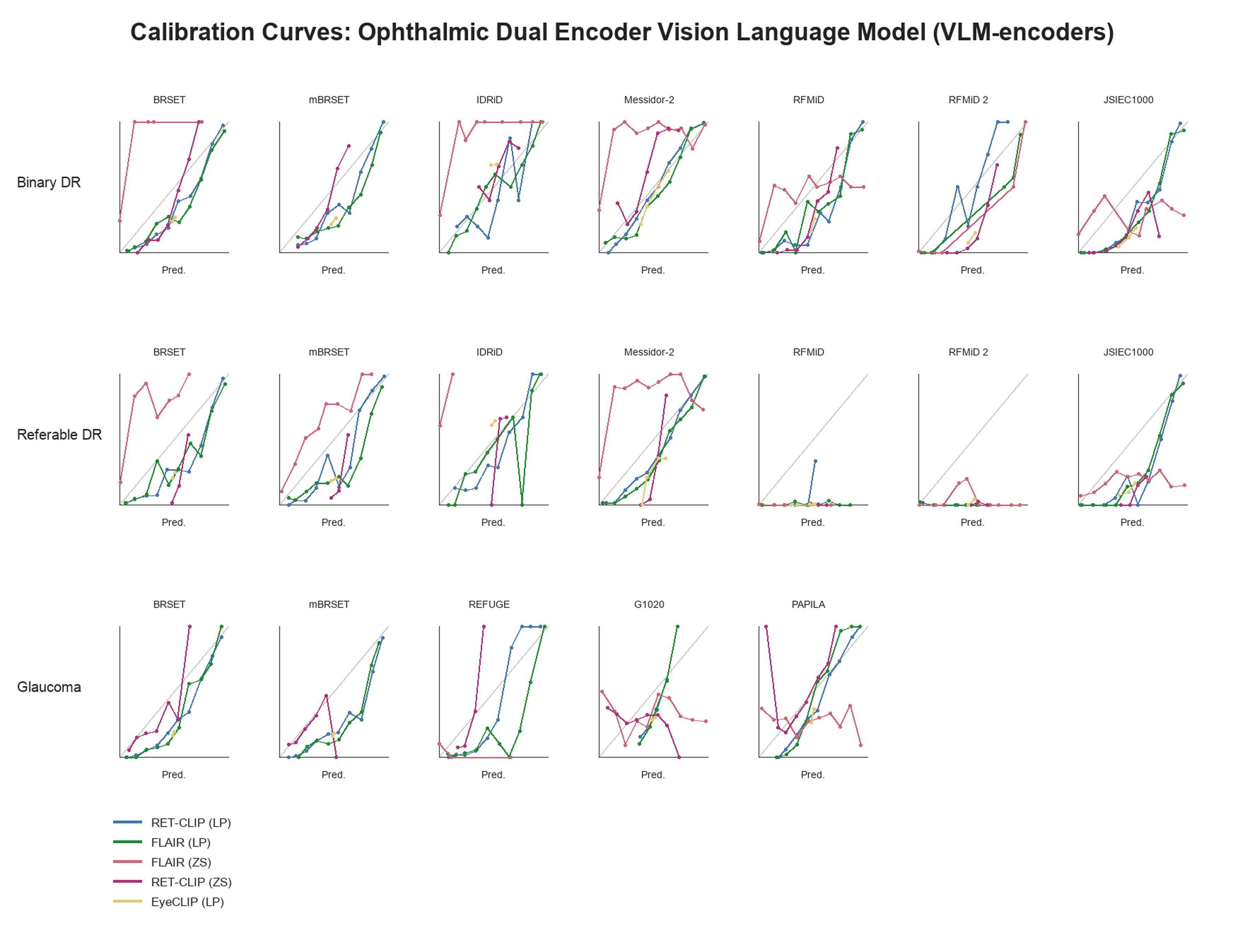}
  \caption{\textbf{Calibration curves for ophthalmic VLMs.} Each panel is a dataset/task pair; the diagonal indicates perfect calibration.}
  \label{fig:supp-calibration-curves-vlm-ophthalmo}
\end{figure}

\begin{figure}[H]
  \centering
  \includegraphics[width=\linewidth]{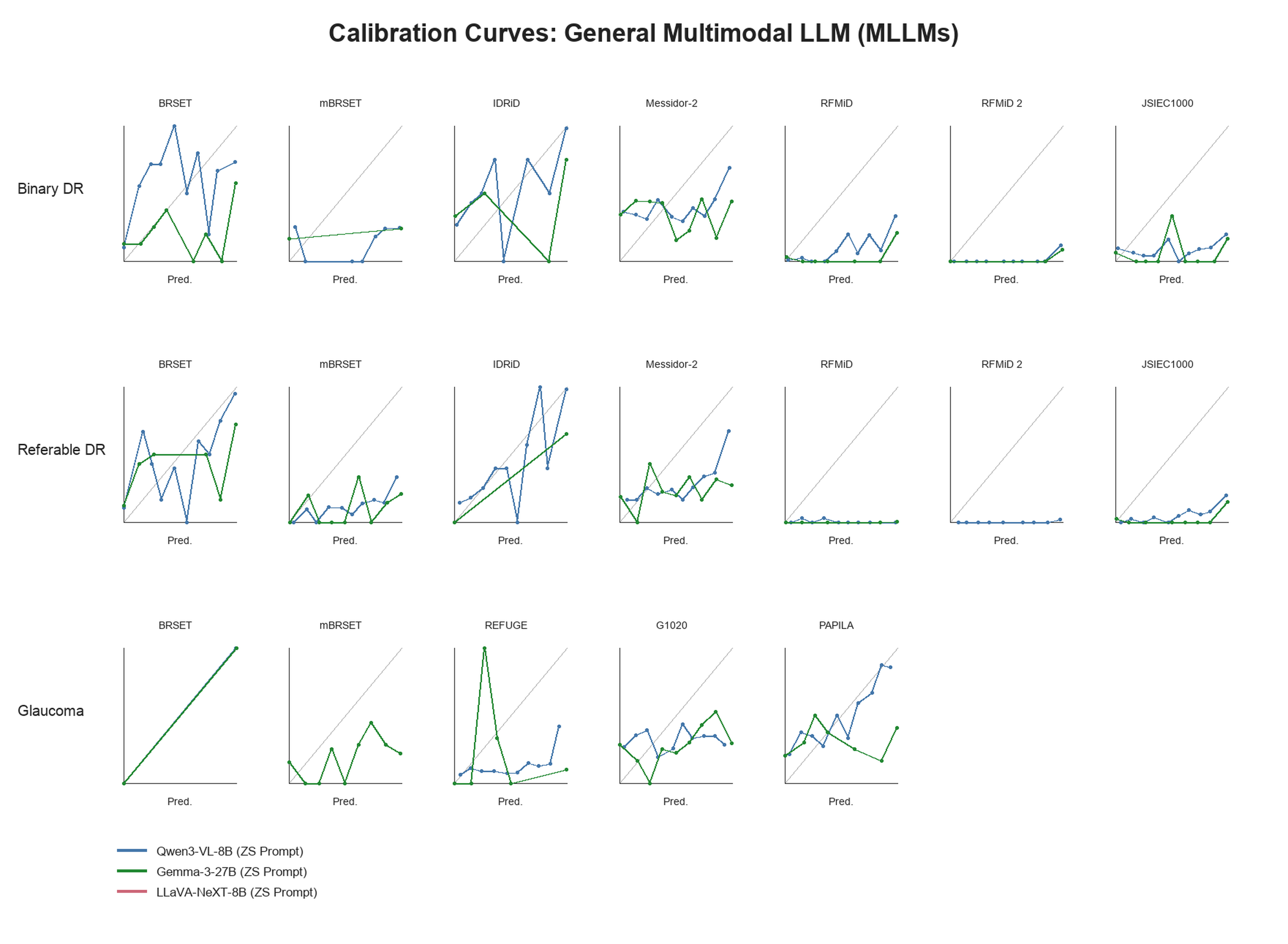}
  \caption{\textbf{Calibration curves for general MLLMs.} Each panel is a dataset/task pair; the diagonal indicates perfect calibration.}
  \label{fig:supp-calibration-curves-mllm-general}
\end{figure}

\begin{figure}[H]
  \centering
  \includegraphics[width=\linewidth]{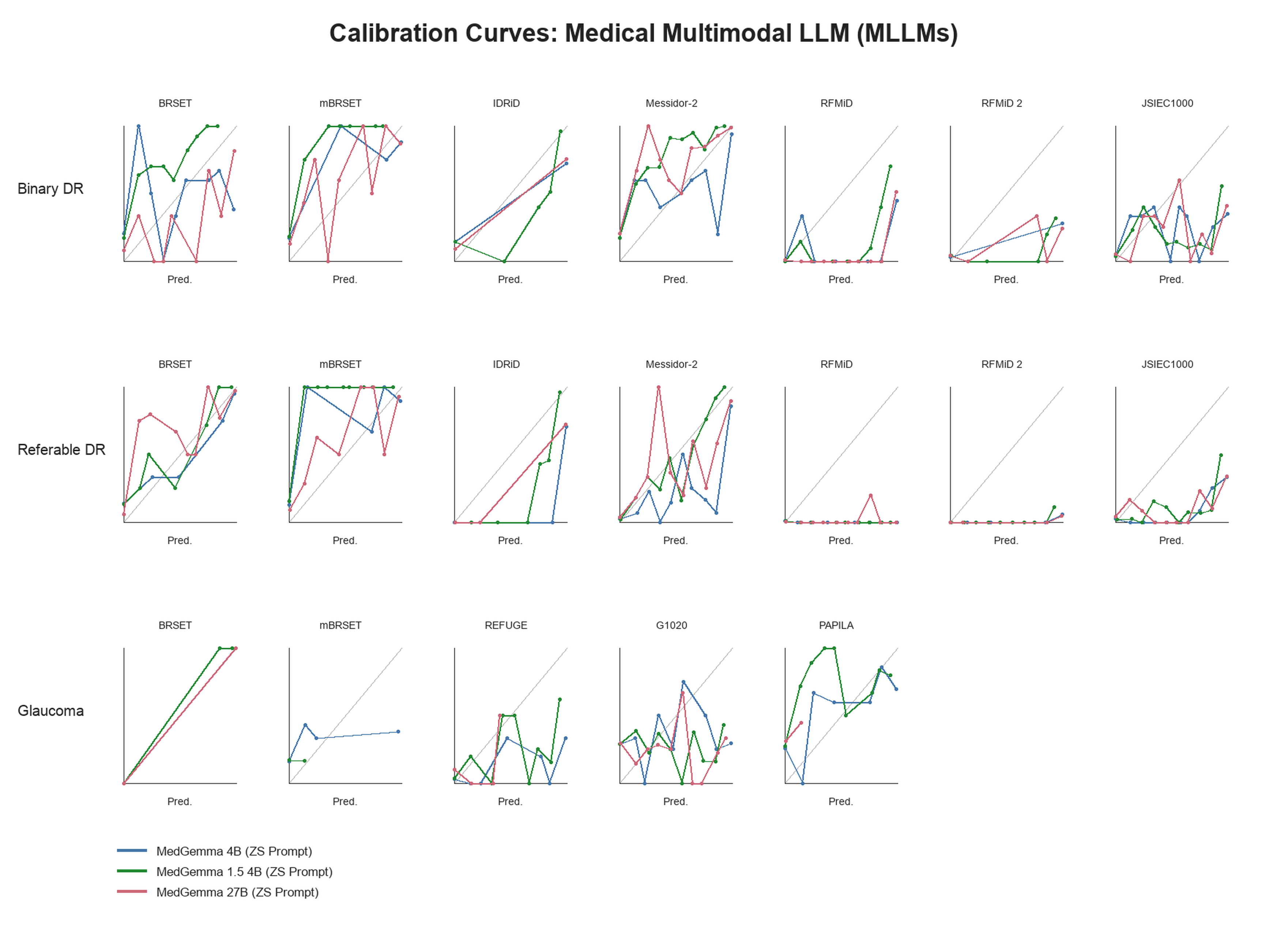}
  \caption{\textbf{Calibration curves for medical MLLMs.} Each panel is a dataset/task pair; the diagonal indicates perfect calibration.}
  \label{fig:supp-calibration-curves-mllm-medical}
\end{figure}

\subsection{Family-specific and adaptation diagnostics}

\begin{figure}[H]
  \centering
  \includegraphics[width=.88\linewidth]{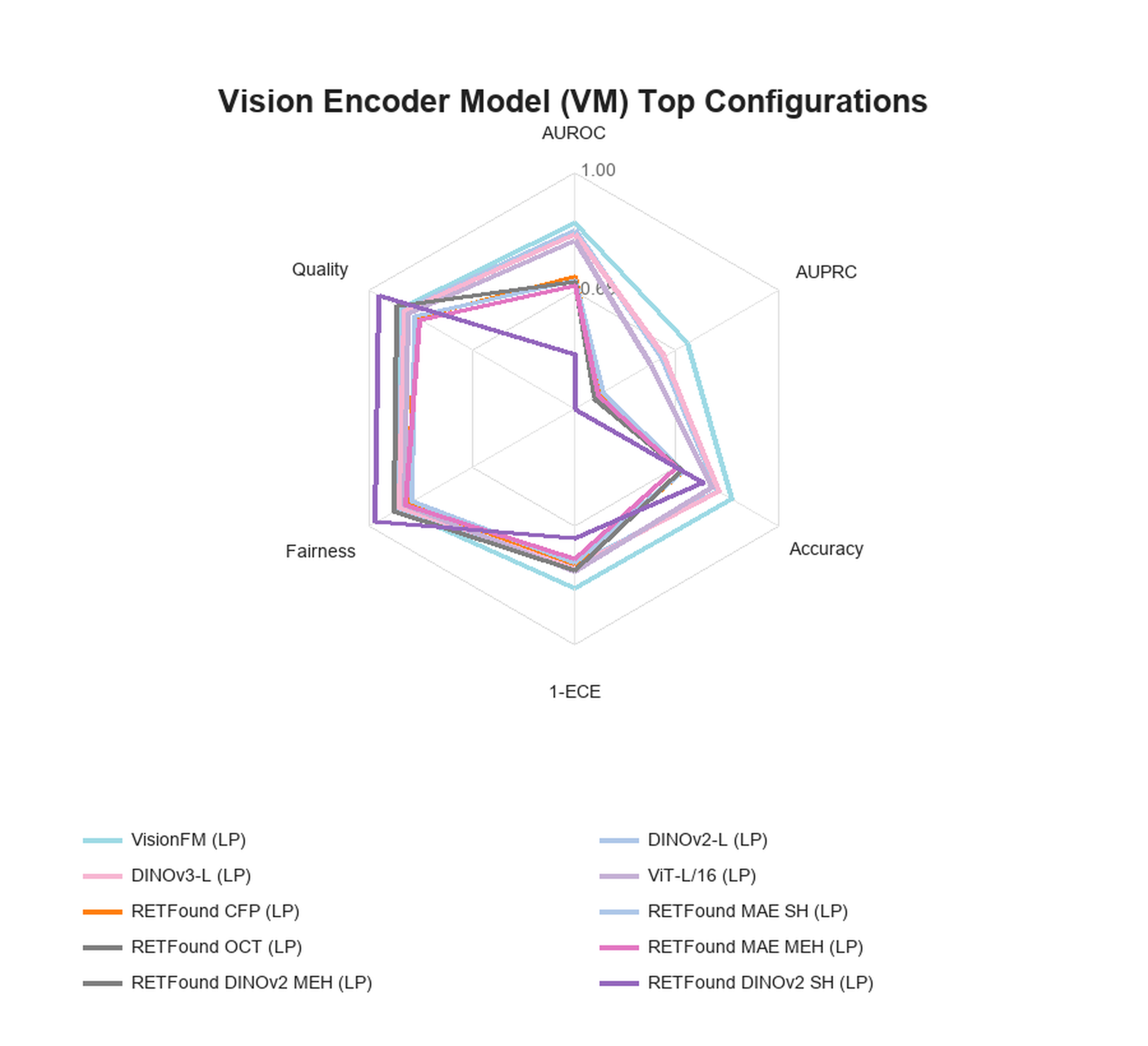}
  \caption{\textbf{VM multimetric spider plot.} Top VM configurations are compared across ranking, thresholded performance, calibration, and available robustness/fairness diagnostics.}
  \label{fig:supp-cv-spider}
\end{figure}

\begin{figure}[H]
  \centering
  \includegraphics[width=.88\linewidth]{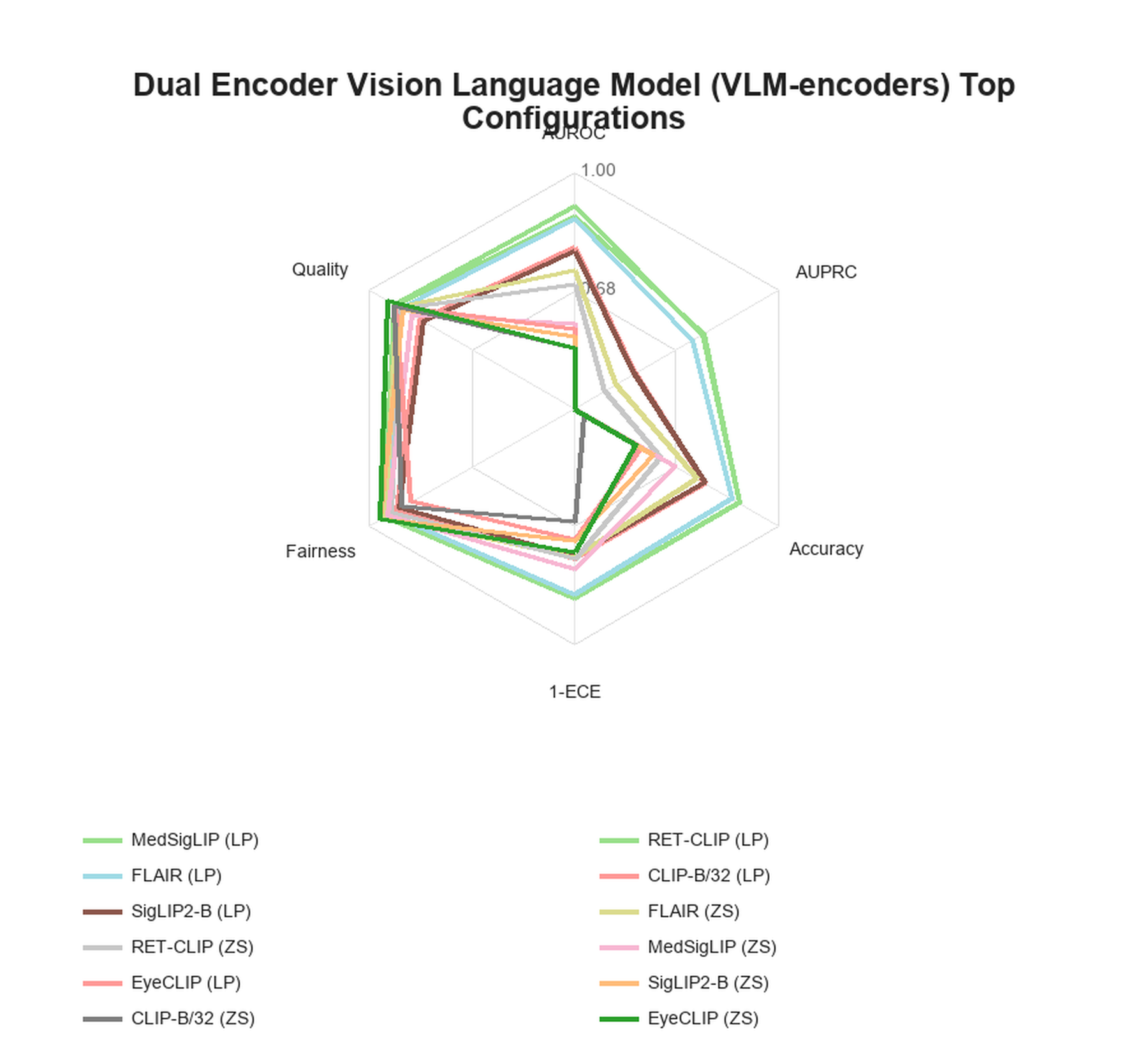}
  \caption{\textbf{VLM multimetric spider plot.} Top VLM configurations show how linear probing changes not only AUROC but also calibration and thresholded behavior.}
  \label{fig:supp-vlm-spider}
\end{figure}

\begin{figure}[H]
  \centering
  \includegraphics[width=.88\linewidth]{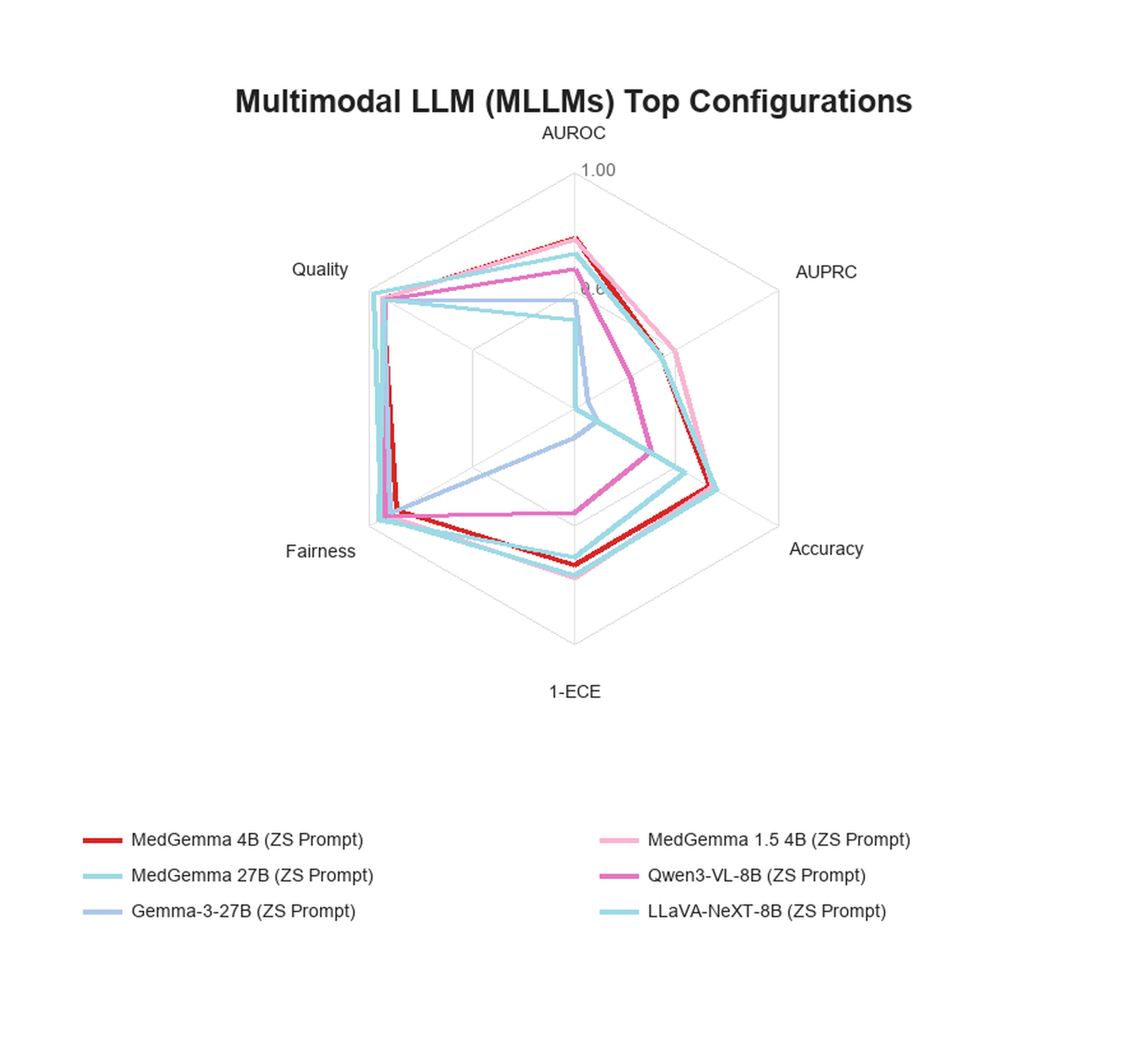}
  \caption{\textbf{MLLM multimetric spider plot.} Top MLLM configurations differ not only in AUROC but also in AUPRC, thresholded accuracy, calibration, and available robustness/fairness diagnostics.}
  \label{fig:supp-mllm-spider}
\end{figure}

\begin{figure}[H]
  \centering
  \includegraphics[width=\linewidth]{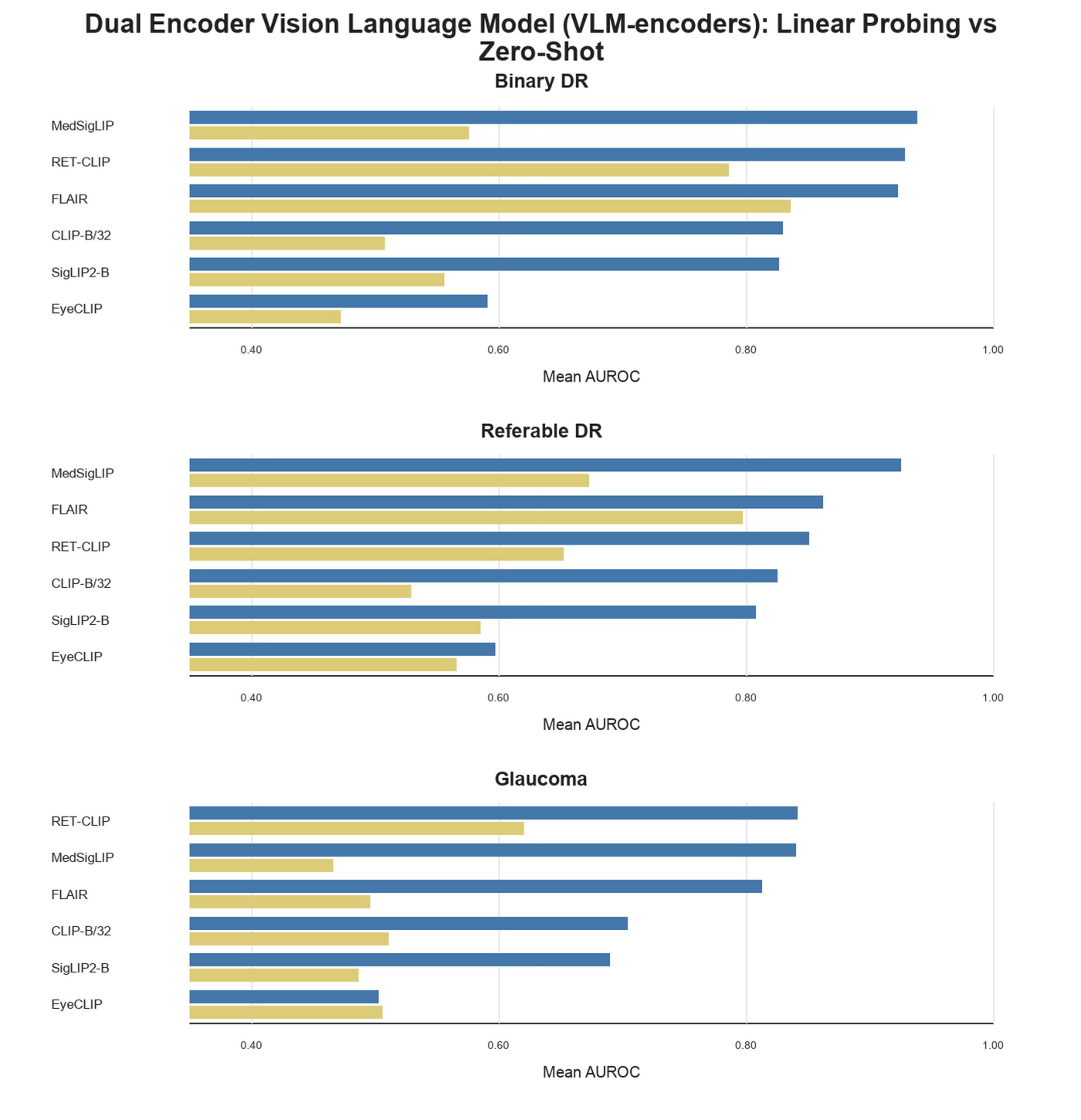}
  \caption{\textbf{VLM zero-shot versus linear-probing comparison.} Linear probing consistently changes the operating point relative to zero-shot image-text matching, motivating method-aware reporting.}
  \label{fig:supp-vlm-adaptation}
\end{figure}

\begin{figure}[H]
  \centering
  \includegraphics[width=\linewidth]{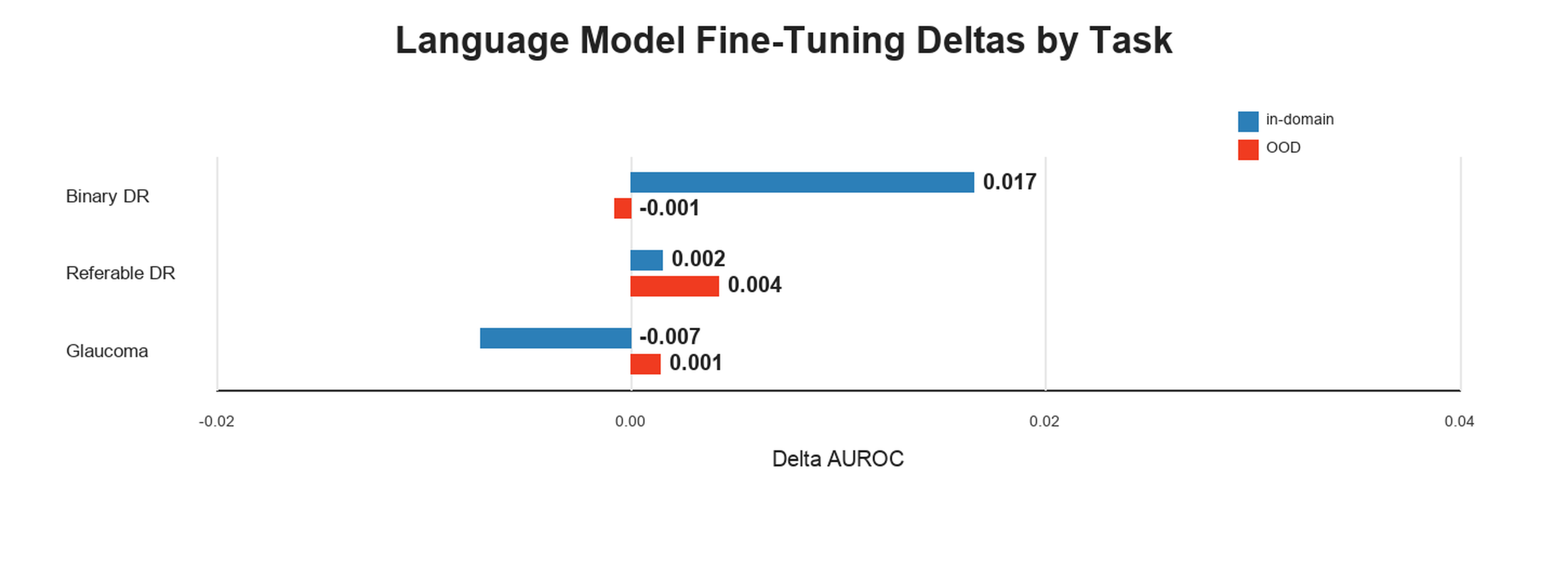}
  \caption{\textbf{Language-model fine-tuning deltas by training task.} SFT effects are not uniform across binary DR, referable DR, and glaucoma.}
  \label{fig:supp-ft-task-deltas}
\end{figure}

\section{Task-Level Comparisons}

\begin{figure}[H]
  \centering
  \includegraphics[width=\linewidth]{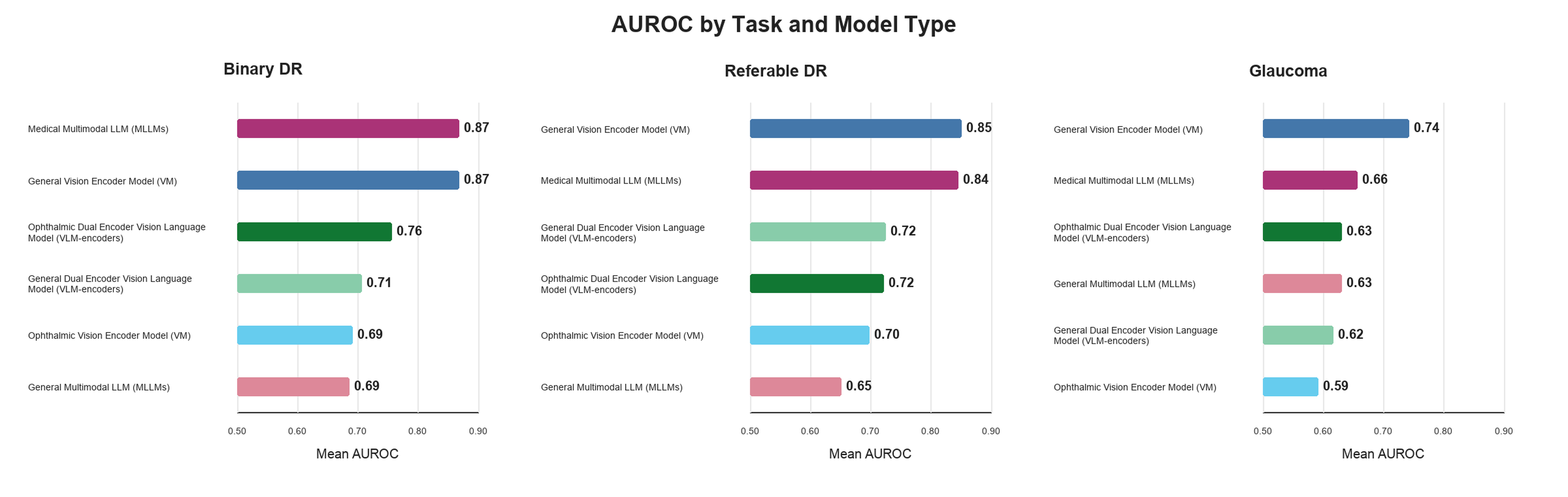}
  \caption{\textbf{Task-level model-type comparison.} Mean AUROC by disease task and model type. Splitting the comparison by task avoids hiding glaucoma-specific weaknesses inside a pooled average.}
  \label{fig:task-model-type}
\end{figure}

\section{Leaderboards}

\begin{figure}[H]
  \centering
  \includegraphics[width=\linewidth]{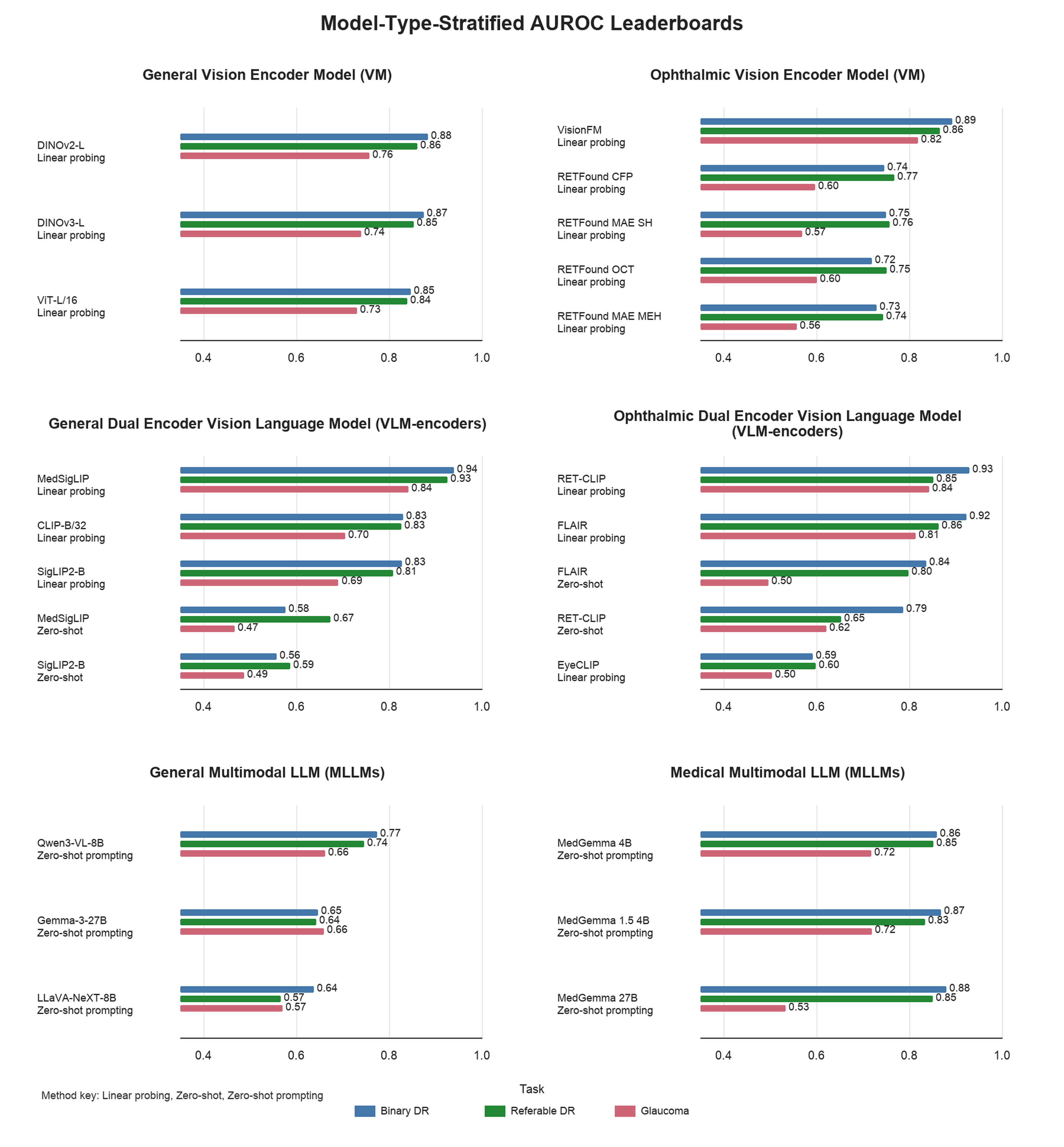}
  \caption{\textbf{Model-type-stratified leaderboards.} Each panel lists the leading model/method configurations within one model type, with separate bars for binary DR, referable DR, and glaucoma. This replaces a single pooled leaderboard so task and model-type effects remain visually distinct.}
  \label{fig:model-leaderboard}
\end{figure}

\section{Interface and Adaptation}

\begin{figure}[H]
  \centering
  \includegraphics[width=\linewidth]{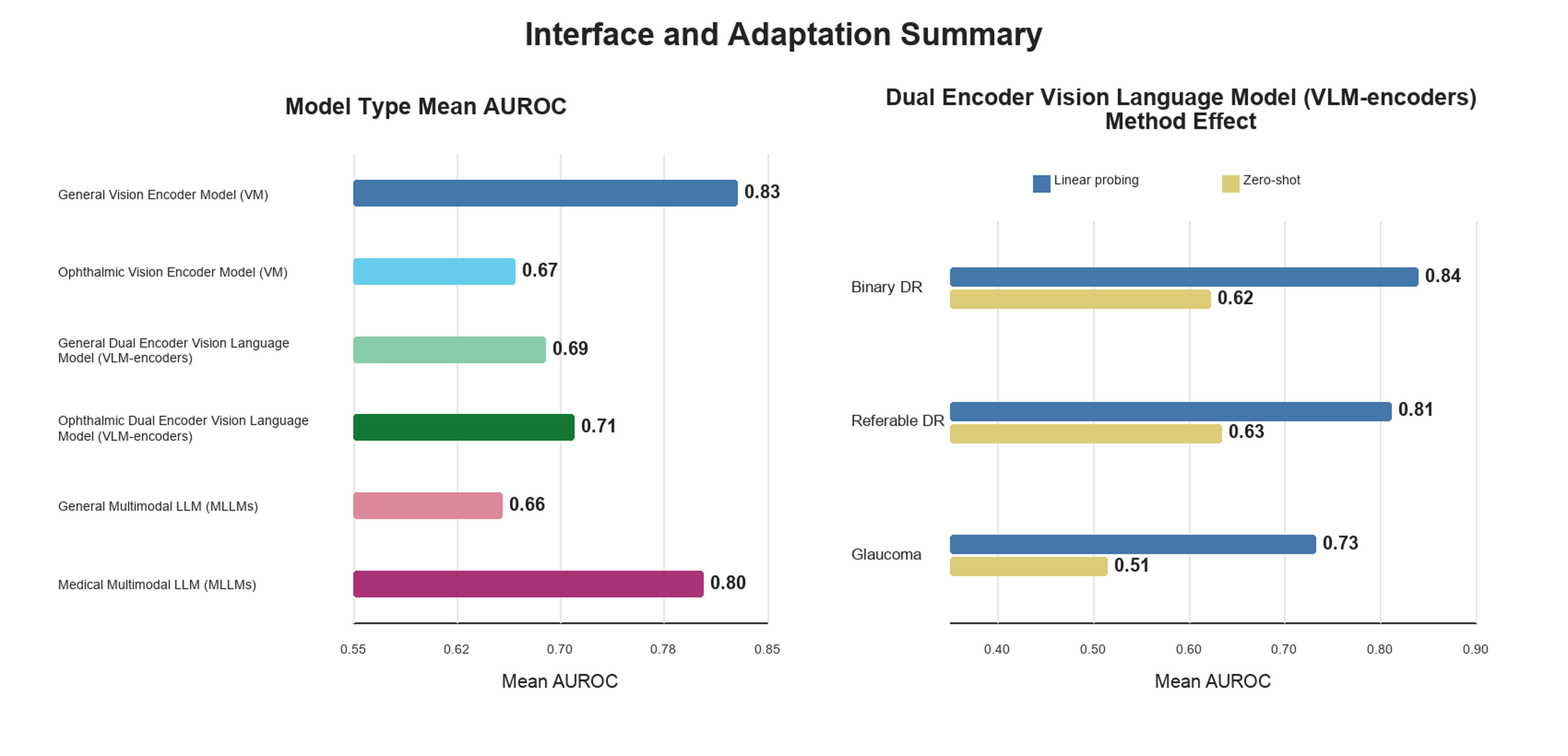}
  \caption{\textbf{Interface and adaptation summary.} Left: mean AUROC by model type. Right: VLM linear probing and zero-shot performance summarized in one task-stratified view.}
  \label{fig:interface-adaptation}
\end{figure}

\section{Reliability Matrix}

\begin{figure}[H]
  \centering
  \includegraphics[width=\linewidth]{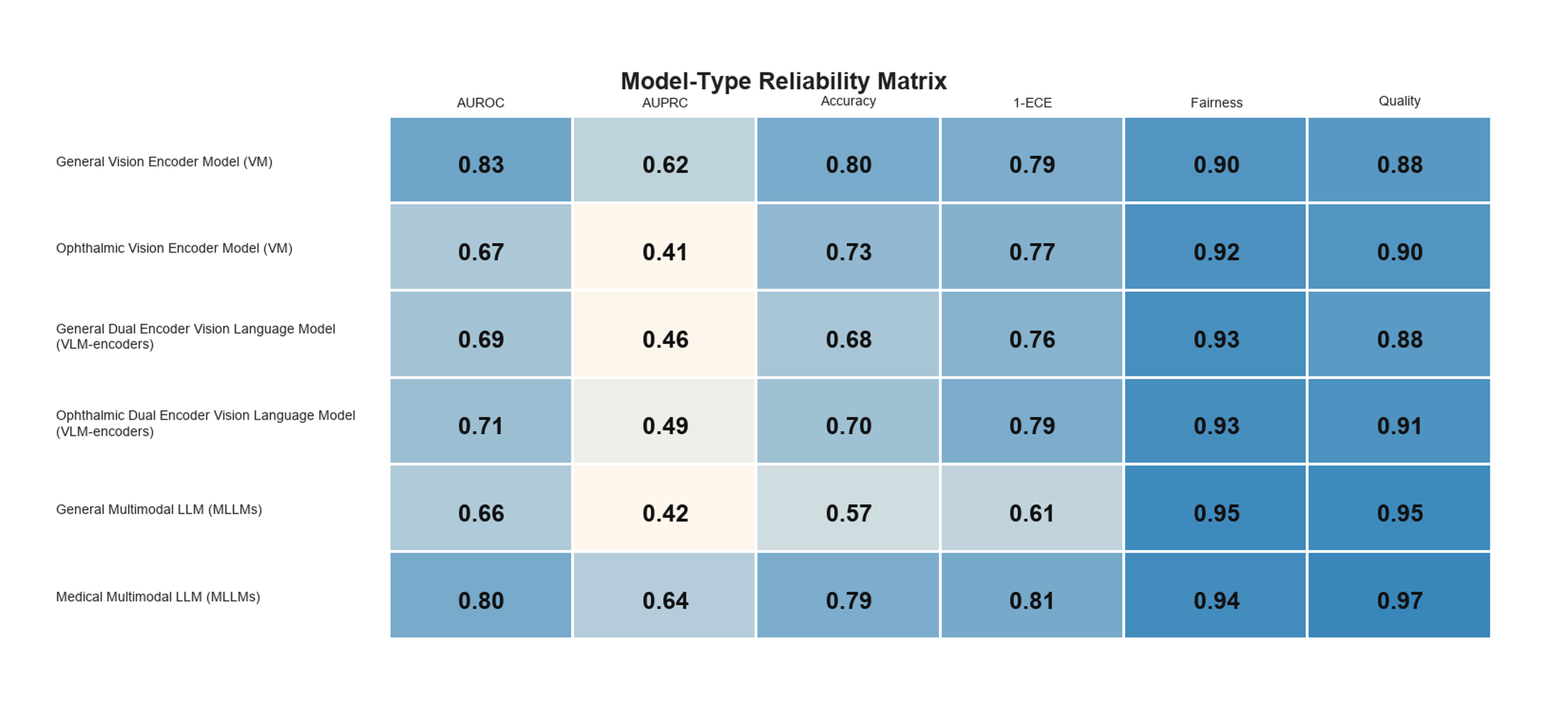}
  \caption{\textbf{Reliability matrix by model type.} Higher values are better. Fairness and quality robustness summarize available metadata-backed gaps, so these cells should be read as coverage-aware diagnostics rather than universal claims for every dataset.}
  \label{fig:reliability-matrix}
\end{figure}

\section{Code and Result Artifacts}

The released repository contains modular dataset loaders, task-label harmonization rules, model-interface wrappers, evaluation routines, and figure-generation utilities. The benchmark artifacts include machine-readable summaries for aggregate performance, calibration, subgroup fairness, image-quality robustness, coverage, and language-model fine-tuning analyses. These artifacts are designed to support both the manuscript figures and the public arena dashboard.

The dataset and model modules are organized as registries rather than one-off analysis code. New datasets can be added by specifying metadata fields, label mappings, supported tasks, and split definitions; new models can be added by implementing the relevant VM, VLM, or MLLM interface. This design keeps the benchmark extensible as additional fundus datasets, acquisition variables, and model families become available.

\section{Dashboard Organization}

The dashboard is organized around the same analysis dimensions used in the benchmark: arena, performance, calibration, fairness, robustness, model types, tasks, coverage, and leaderboard tables. Fine-tuned SFT adapter analyses are presented as a separate adaptation view so that base benchmark results are not mixed with fine-tuning experiments.

\section{Ethics and Limitations}

\benchmark{} is retrospective and should not be interpreted as clinical validation. Labels may be noisy or non-equivalent across sources. Demographic and acquisition metadata are incomplete. Subgroup gaps are descriptive and may be unstable for small strata. Dataset licenses may restrict image redistribution. Public dashboards should therefore release metrics, metadata, manifests, and code only where permitted.


\end{document}